\documentclass{article} 
\usepackage{iclr_r2fm,times}

\usepackage{amsmath,amsfonts,bm, bbm}
\usepackage{xparse}

\DeclareDocumentCommand\W{ g g }{%
        \IfNoValueTF {#1} {\mathbf{W}} {
            \IfNoValueTF {#2} {\mathbf{W}^{(#1)}}{\mathbf{W}^{(#1)}_{#2}}
        }
}

\DeclareDocumentCommand\bias{ g g }{%
        \IfNoValueTF {#1} {\mathbf{b}} {
            \IfNoValueTF {#2} {\mathbf{b}^{(#1)}}{\mathbf{b}^{(#1)}_{#2}}
        }
}

\DeclareDocumentCommand\betavar{ g g }{%
        \IfNoValueTF {#1} {\bm{\beta}} {
            \IfNoValueTF {#2} {{\bm{\beta}^{(#1)}}{}}{\bm{\beta}^{(#1)}_{#2}}
        }
}

\DeclareDocumentCommand\xivar{ g g }{%
        \IfNoValueTF {#1} {\bm{\xi}} {
            \IfNoValueTF {#2} {{\bm{\xi}^{(#1)}}{}}{\bm{\xi}^{(#1)}_{#2}}
        }
}

\DeclareDocumentCommand\xivarn{ g g }{%
        \IfNoValueTF {#1} {\bm{\xi^-}} {
            \IfNoValueTF {#2} {\bm{\xi^-}^{+(#1)}}{\bm{\xi^-}^{+(#1)}_{#2}}
        }
}

\DeclareDocumentCommand\xivarp{ g g }{%
        \IfNoValueTF {#1} {\bm{\xi^+}} {
            \IfNoValueTF {#2} {\bm{\xi^+}^{+(#1)}}{\bm{\xi^+}^{+(#1)}_{#2}}
        }
}

\DeclareDocumentCommand\nuvar{ g g }{%
        \IfNoValueTF {#1} {\bm{\nu}} {
            \IfNoValueTF {#2} {{\bm{\nu}^{(#1)}}{}}{\bm{\nu}^{(#1)}_{#2}{}}
        }
}

\DeclareDocumentCommand\hnuvar{ g g }{%
        \IfNoValueTF {#1} {\bm{\hat{\nu}}} {
            \IfNoValueTF {#2} {{\bm{\hat{\nu}}^{(#1)}}{}}{\bm{\hat{\nu}}^{(#1)}_{#2}{}}
        }
}

\DeclareDocumentCommand\muvar{ g g }{%
        \IfNoValueTF {#1} {\bm{\mu}} {
            \IfNoValueTF {#2} {{\bm{\mu}^{(#1)}}{}}{\bm{\mu}^{(#1)}_{#2}}
        }
}

\DeclareDocumentCommand\gammavar{ g g }{%
        \IfNoValueTF {#1} {\bm{\gamma}} {
            \IfNoValueTF {#2} {{\bm{\gamma}^{(#1)}}{}}{\bm{\gamma}^{(#1)}_{#2}}
        }
}

\DeclareDocumentCommand\lambdavar{ g g }{%
        \IfNoValueTF {#1} {\bm{\lambda}} {
            \IfNoValueTF {#2} {{\bm{\lambda}^{(#1)}}{}}{\bm{\lambda}^{(#1)}_{#2}}
        }
}

\DeclareDocumentCommand\tbetavar{ g g }{%
        \IfNoValueTF {#1} {{\bm{\tilde{\beta}}}} {
            \IfNoValueTF {#2} {{{\bm{\tilde{\beta}}}^{(#1)}}{}}{{{\bm{\tilde{\beta}}}^{(#1)}_{#2}}}
        }
}

\DeclareDocumentCommand\alphavar{ g g }{%
        \IfNoValueTF {#1} {\bm{\alpha}} {
            \IfNoValueTF {#2} {{\bm{\alpha}^{(#1)}}}{\bm{\alpha}^{(#1)}_{#2}}
        }
}

\DeclareDocumentCommand\D{ g g }{%
        \IfNoValueTF {#1} {\mathbf{D}} {
            \IfNoValueTF {#2} {\mathbf{D}^{(#1)}}{\mathbf{D}^{(#1)}_{#2}}
        }
}

\DeclareDocumentCommand\A{ g g }{%
        \IfNoValueTF {#1} {\mathbf{A}} {
            \IfNoValueTF {#2} {\mathbf{A}^{(#1)}}{\mathbf{A}^{(#1)}_{#2}}
        }
}

\DeclareDocumentCommand\AA{ g g }{
        \IfNoValueTF {#1} {\mathbf{\Omega}} {
            \IfNoValueTF {#2} {\mathbf{\Omega}(#1, #1)}{\mathbf{\Omega}(#1, #2)}
        }
}

\DeclareDocumentCommand\S{ g g }{%
        \IfNoValueTF {#1} {\mathbf{S}} {
            \IfNoValueTF {#2} {\mathbf{S}^{(#1)}}{\mathbf{S}^{(#1)}_{#2}}
        }
}

\DeclareDocumentCommand\K{ g g }{%
        \IfNoValueTF {#1} {\mathbf{K}} {
            \IfNoValueTF {#2} {\mathbf{K}^{(#1)}}{\mathbf{K}^{(#1)}_{#2}}
        }
}

\DeclareDocumentCommand\B{ g g }{%
        \IfNoValueTF {#1} {\mathbf{B}} {
            \IfNoValueTF {#2} {\mathbf{B}^{(#1)}}{\mathbf{B}^{(#1)}_{#2}}
        }
}

\DeclareDocumentCommand\lowerb{ g g }{%
        \IfNoValueTF {#1} {{\mathbf{\underline{b}}}} {
            \IfNoValueTF {#2} {{\mathbf{\underline{b}}}^{(#1)}}{{\mathbf{\underline{b}}}^{(#1)}_{#2}}
        }
}

\DeclareDocumentCommand\z{ g g }{%
        \IfNoValueTF {#1} {z} {
            \IfNoValueTF {#2} {z^{(#1)}}{z^{(#1)}_{#2}}
        }
}

\DeclareDocumentCommand\s{ g g }{%
        \IfNoValueTF {#1} {s} {
            \IfNoValueTF {#2} {s^{(#1)}}{s^{(#1)}_{#2}}
        }
}

\DeclareDocumentCommand\dom{ g g }{%
        \IfNoValueTF {#1} {\mathcal{S}} {
            \IfNoValueTF {#2} {\mathcal{S}_{#1}}{\mathcal{S}^{#1}_{#2}}
        }
}

\DeclareDocumentCommand\domlb{ g g }{%
        \IfNoValueTF {#1} {\mathsf{LB}} {
            \IfNoValueTF {#2} {\mathsf{LB}(\mathcal{#1})}{\mathsf{LB}(\mathcal{#1}_{#2})}
        }
}

\DeclareDocumentCommand\domub{ g g }{%
        \IfNoValueTF {#1} {\mathsf{UB}} {
            \IfNoValueTF {#2} {\mathsf{UB}(\mathcal{#1})}{\mathsf{UB}(\mathcal{#1}_{#2})}
        }
}

\DeclareDocumentCommand\uns{ g g }{%
        \IfNoValueTF {#1} {\tilde{s}} {
            \IfNoValueTF {#2} {\tilde{s}_{#1}}{s^{(#1)}_{#2}}
        }
}

\DeclareDocumentCommand\ub{ g g }{%
        \IfNoValueTF {#1} {u} {
            \IfNoValueTF {#2} {u^{(#1)}}{u^{(#1)}_{#2}}
        }
}

\DeclareDocumentCommand\lb{ g g }{%
        \IfNoValueTF {#1} {l} {
            \IfNoValueTF {#2} {l^{(#1)}}{l^{(#1)}_{#2}}
        }
}

\DeclareDocumentCommand\hz{ g g }{%
        \IfNoValueTF {#1} {\hat{z}} {
            \IfNoValueTF {#2} {\hat{z}^{(#1)}}{\hat{z}^{(#1)}_{#2}}
        }
}

\DeclareDocumentCommand\bu{ g g }{%
        \IfNoValueTF {#1} {\mathbf{u}} {
            \IfNoValueTF {#2} {\mathbf{u}^{(#1)}}{\mathbf{u}^{(#1)}_{#2}}
        }
}

\DeclareDocumentCommand\bl{ g g }{%
        \IfNoValueTF {#1} {\mathbf{l}} {
            \IfNoValueTF {#2} {\mathbf{l}^{(#1)}}{\mathbf{l}^{(#1)}_{#2}}
        }
}

\DeclareDocumentCommand\aaa{ g }{%
        \IfNoValueTF {#1} {\bm{a}} {
            {\bm{a}^{({#1})}}
        }
}

\DeclareDocumentCommand\haaa{ g }{%
        \IfNoValueTF {#1} {\bm{\hat{a}}} {
            {\bm{\hat{a}}^{({#1})}}
        }
}

\DeclareDocumentCommand\bbb{ g g }{%
        \IfNoValueTF {#1} {\mathbf{P}} {
            \IfNoValueTF {#2} {{\mathbf{P}_{#1}}}{{\mathbf{P}_{#1}^{({#2})}}}
        }
}

\DeclareDocumentCommand\hbbb{ g g }{%
        \IfNoValueTF {#1} {\mathbf{\hat{P}}} {
            \IfNoValueTF {#2} {{\mathbf{\hat{P}}_{#1}}}{{\mathbf{\hat{P}}_{#1}^{({#2})}}}
        }
}

\DeclareDocumentCommand\ccc{ g g }{%
        \IfNoValueTF {#1} {\mathbf{q}} {
            \IfNoValueTF {#2} {{\mathbf{q}_{#1}}}{{\mathbf{q}_{#1}^{(#2)}}{}}
        }
}

\DeclareDocumentCommand\constc{ g }{%
        \IfNoValueTF {#1} {c} {
            {c^{({#1})}}
        }
}

\DeclareDocumentCommand\setz{ g g }{%
        \IfNoValueTF {#1} {\mathcal{Z}} {
            \IfNoValueTF {#2} {\mathcal{Z}^{(#1)}}{\mathcal{Z}^{(#1)}_{#2}}
        }
}

\DeclareDocumentCommand\setzp{ g g }{%
        \IfNoValueTF {#1} {\mathcal{Z^+}} {
            \IfNoValueTF {#2} {\mathcal{Z}^{+(#1)}}{\mathcal{Z}^{+(#1)}_{#2}}
        }
}

\DeclareDocumentCommand\setzn{ g g }{%
        \IfNoValueTF {#1} {\mathcal{Z^-}} {
            \IfNoValueTF {#2} {\mathcal{Z}^{-(#1)}}{\mathcal{Z}^{-(#1)}_{#2}}
        }
}

\DeclareDocumentCommand\tsetz{ g g }{%
        \IfNoValueTF {#1} {\tilde{\mathcal{Z}}} {
            \IfNoValueTF {#2} {\tilde{\mathcal{Z}}^{(#1)}}{\tilde{\mathcal{Z}}^{(#1)}_{#2}}
        }
}

\DeclareDocumentCommand\tz{ g g }{%
        \IfNoValueTF {#1} {\tilde{z}} {
            \IfNoValueTF {#2} {\tilde{z}^{(#1)}}{\tilde{z}^{(#1)}_{#2}}
        }
}

\DeclareDocumentCommand\f{ g g }{%
        \IfNoValueTF {#1} {f} {
            \IfNoValueTF {#2} {f^{(#1)}}{f^{(#1)}_{#2}}
        }
}

\DeclareDocumentCommand\lf{ g g }{%
        \IfNoValueTF {#1} {\underline{f}} {
            \IfNoValueTF {#2} {\underline{f}^{(#1)}}{\underline{f}^{(#1)}_{#2}}
        }
}

\def\eqref#1{Eq.~(\ref{#1})}

\def\1{\bm{1}}

\def\vzero{{\bm{0}}}

\def\vtheta{{\bm{\theta}}}
\def\vepsilon{{\bm{\epsilon}}}

\def\vI{{\bm{I}}}
\def\vdelta{{\bm{\delta}}}

\def\vx{{\bm{x}}}
\def\vy{{\bm{y}}}
\def\vz{{\bm{z}}}

\DeclareMathAlphabet{\mathsfit}{\encodingdefault}{\sfdefault}{m}{sl}
\SetMathAlphabet{\mathsfit}{bold}{\encodingdefault}{\sfdefault}{bx}{n}

\usepackage{hyperref}
\usepackage{url}
\usepackage{amsmath}
\usepackage{ragged2e} 
\usepackage{booktabs,makecell, multirow, tabularx}
\usepackage{subfigure}
\usepackage{wrapfig}
\usepackage{adjustbox}
\usepackage{caption}
\usepackage{CJK}
\usepackage[noend]{algpseudocode}
\usepackage{algorithmicx,algorithm}
\counterwithin{section}{part}
\renewcommand{\thesection}{\arabic{section}}

\title{Unlearnable Examples for Diffusion Models:\\ Protect Data from Unauthorized Exploitation}

\author{Zhengyue Zhao $^{1,2}$, Jinhao Duan $^3$, Xing Hu $^{1,4}$\thanks{ Corresponding author.}, Kaidi Xu $^3$, Chenan Wang $^3$ \\
\textbf{Rui Zhang $^{1}$, Zidong Du $^{1,4}$, Qi Guo$^{1}$, Yunji Chen $^{1,2}$}  \\
$^1$ SKL of Processors, Institute of Computing Technology, Chinese Academy of Sciences \\
$^2$ University of Chinese Academy of Sciences\\
$^3$ Drexel University\\ $^4$ Shanghai Innovation Center for Processor Technologies, SHIC \\
\texttt{zhaozhengyue22@mails.ucas.ac.cn} \\
\texttt{\{jd3734, kx46, cw3344\}@drexel.edu} \\
\texttt{\{huxing, zhangrui, duzidong, guoqi, cyj\}@ict.ac.cn}
}

\iclrfinalcopy 
\begin{document}

\maketitle

\begin{abstract}
  Diffusion models have demonstrated remarkable performance in image generation tasks while also raising security and privacy concerns. To tackle these issues, we propose a method for generating unlearnable examples for diffusion models, Unlearnable Diffusion Perturbation, to safeguard images from unauthorized exploitation. Our approach involves designing an algorithm to generate sample-wise perturbation noise for each image to be protected. We frame this as a max-min optimization problem and introduce EUDP, a noise scheduler-based method to enhance the effectiveness of the protective noise. Our experiments demonstrate that training diffusion models on the protected data leads to a significant reduction in the quality of the generated images.
\end{abstract}

\section{Introduction}

In recent years, generative models such as GANs~\citep{goodfellow2020generative,brock2018large} and VAEs~\citep{kingma2013auto} have made significant strides in image synthesis tasks. Notably, Denoising Diffusion Probabilistic Model (DDPM)~\citep{sohl2015deep,ho2020denoising} and other diffusion models~\citep{nichol2021improved,song2020score,bao2022analytic,song2020score} have surpassed GANs in performance~\citep{dhariwal2021diffusion}, becoming the state-of-the-art image synthesis method. Furthermore, the use of Latent Diffusion Models (LDM)~\citep{rombach2022high} has enhanced the ability to generate high-resolution images and perform multi-model tasks, such as text-to-image generation, leading to diverse AI-for-Art applications such as Stable Diffusion and MidJourney. While training a high-performance diffusion model from scratch remains a costly endeavor, there are effective fine-tuning techniques such as Textual Inversion~\citep{gal2022image} and DreamBooth~\citep{ruiz2022dreambooth} that allow for personalized diffusion models to be trained from large pre-trained models with minimal training overhead and small datasets.

However, the development of diffusion-based image synthesis methods has also given rise to security and privacy concerns. Abused unauthorized data exploitation is one major issue for generative models. For example, a pre-trained diffusion model can be fine-tuned with several personal facial images to generate fake images that could be harmful to the owner. Additionally, using artworks for training diffusion models could result in copyright infringement, dampening artists' creative enthusiasm. While artists may want to share their work on social networks, they may not want their work to be used for unauthorized exploitation, such as training a diffusion model~\citep{bbc2022Art,cnn2022AI,washington2022AI}. Therefore, it is imperative to protect artworks without hindering their normal usage. Unlearnable  examples~\citep{huang2021unlearnable, fu2022robust}  are firstly proposed to protect data from being unauthorizedly trained by neural models for image classification via adding imperceptible perturbation to original images, which effectively safeguards the privacy and copyright of personal data.

Motivated by unlearnable examples for image classification, we adapt unlearnable examples to diffusion models in this paper and propose Unlearnable Diffusion Perturbation (UDP) designed to protect images from being utilized to train a high-performance diffusion model by adding perturbation protective noise to images. The protected images have minimal differences compared to the original images but the diffusion model trained on these protected images is unable to generate the expected high-quality images. Additionally, we observe that modifying the sampling scheme of timesteps during the optimization process based on the noise scheduler of the diffusion process can further strengthen the protective effect of the noise. Building upon this observation, we introduce an improved method, Enhanced Unlearnable Diffusion Perturbation (EUDP) to further enhance the protection.

\section{Problem Formalization}
We formalize the normal diffusion model training and the protective unlearnable noise generation process as follows.

\textbf{Normal diffusion model training:} For a clean training image dataset $\vx \in S_c$, which follows the distribution $\vx\sim q^c(\vx)$, an image generator model $G_\vtheta(\vx)$ is trained to generate images following the distribution $\hat{\vx}\sim p^c_\vtheta(\vx)$ that is as close as possible to $q^c(\vx)$. 
The distance between these two distributions $D(p_\vtheta^c(\vx),q^c(\vx))$ can be evaluated by using some distance metrics, such as KL divergence.

\textbf{Protective Noise Generation:}
By adding a small amount of protective noise $\vdelta^u$ (bounded by $\rho_u$) to the clean training dataset, we obtain unlearnable examples $\vx^u=\vx+\vdelta^u$, where $\vx^u \in S_u$ and follow the distribution $\vx^u\sim q^u(\vx)$. For the image generator model $G_\vtheta(\vx^u)$ trained on unlearnable data, the generated images follow the distribution $\hat{\vx}^u\sim p_\vtheta^u(\vx)$. 
\begin{equation}
\label{equ:max_dist}
\max_{\Vert\vdelta^u\Vert\leq\rho_u}D\left(p^u_\vtheta(\vx),q^c(\vx)\right)
\end{equation}
To achieve the unlearnable effect, the design objective is to increase the distance in Eq.~\ref{equ:max_dist} between the distribution of generated images and the distribution of clean training data as much as possible by adjusting the protective noise $\vdelta^u$.

\section{Methodology}
\subsection{Unlearnable Diffusion Perturbation}

We propose the Unleanable Difussion Perturbation (UDP) method for protective noise generation. For DDPM, the training images follow the distribution $q(\vx)$, while the generated images follow the distribution $p_\vtheta (\vx)$, with the optimization objective being the cross-entropy between these two distributions:
\begin{equation}
\label{equ:min_Lce}
    \min_\vtheta \mathcal{L}_{\text{CE}}=\min_\vtheta \mathbb{E}_{q(\vx_0)}\left[-\log{p_\vtheta(\vx_0)}\right]\leq \min_\vtheta \mathcal{L}_{\text{VLB}}=\min_\vtheta \mathbb{E}_{q(\vx_{0:T})}\left[\log{ \frac{q(\vx_{1:T}|\vx_0)}{p_\vtheta (\vx_{0:T})}}\right]
\end{equation}
To address the optimization objective in Eq.~\ref{equ:max_dist}, we approximate the optimization of the cross-entropy by optimizing the variational bound. To reduce the image generation quality of DDPM, we achieve this by maximizing the loss function during the training process. Similarly, we transform the problem of maximizing the optimal value of the cross-entropy in Eq.~\ref{equ:min_Lce} into finding the maximum value of the minimum of the variational bound:
\begin{equation}
\label{equ:maxmin_Lce}
\begin{split}
    \max_\vdelta\min_\vtheta \mathcal{L}_{\text{CE}}&=\max_\vdelta\min_\vtheta \mathbb{E}_{q_\vdelta(\vx_0)}\left[-\log{p_{\vtheta,\vdelta}(\vx_0)}\right] \\ 
    &\leq \max_\vdelta\min_\vtheta \mathcal{L}_{\text{VLB}}=\max_\vdelta\min_\vtheta \mathbb{E}_{q_\vdelta(\vx_{0:T})}\left[\log{\frac{q_\vdelta(\vx_{1:T}|\vx_0)}{p_{\vtheta,\vdelta} (\vx_{0:T})}}\right]
\end{split}
\end{equation}
Similar to the training of DDPM, we expand the terms of $\mathcal{L}_{\text{VLB}}$ in Eq.~\ref{equ:maxmin_Lce}  and replace it with the simplified loss function $\mathcal{L}_{\text{simple}}$ in Eq.~\ref{equ:dm_loss_simple}. Then the optimization objective is to solve the following bi-level max-min optimization problem:
\begin{equation}
\label{equ:UDP_maxmin}
    \vdelta^u=\arg \max_{\Vert\vdelta^u_i\Vert\leq\rho_u} \min_\vtheta \left( \sum_{\vx_i} {\mathbb{E}_{\vx_{i,t_n}\sim p_\vtheta(\vx), t_n\sim \mathcal{U}(0,T)} \mathcal{L}_{t_n}\left(f_\vtheta(\vx_{i,t_n}+\vdelta^u_i)\right)} \right)
\end{equation}
Where $\vdelta^u$ represents the protective noise added to the training data $\vx$, and its $L_\infty$ norm is constrained by the preset value $\rho_u$ to limit the impact of the protective noise on the original image. Considering that direct solution to the max-min problem in Eq.~\ref{equ:UDP_maxmin} is difficult, an iterative method of updating $\vtheta$ and $\vdelta^u_i$ alternatively can be used for optimization:
\begin{equation}
\label{equ:UDP_iter}
\begin{split}
    &\vtheta \gets \vtheta-\eta \cdot \nabla_\vx \mathcal{L}_{t_n}\left( f_\vtheta(\vx_{i,t_n}+\vdelta^u_i) \right)  \\
    &\vdelta^u_i \gets \vdelta^u_i+\lambda \cdot \text{sign} \left( \nabla_\vx \mathcal{L}_{t_n}\left( f_\vtheta(\vx_{i,t_n}+\vdelta^u_i) \right)\right)
\end{split}
\end{equation}

Where $\vx_i$ is randomly sampled from the training data and $t$ is randomly sampled following uniform distribution $\mathcal{U}(0,T)$. The implementation of UDP is demonstrated in Appendix ~\ref{sec:Implementation of UDP}.

\subsection{Enhanced Unlearnable Diffusion Perturbation}
\label{sec:EUDP}
Considering multiple iterative steps of DDPM in its forward process, we make the following two observations to help further improve the protective effect of UDP.

\paragraph{Observation 1: Decay Effect of Protective Noises} Though all the iterative steps ($t$ from $0$ to $T$) are considered when computing protective noise, the protective noise can only be added to the image in the first step ($t=0$). This causes the protective noise to be gradually overwhelmed in the forward process of DDPM:
\begin{equation}
\label{equ:perturb_decay}
    \vx_t+\vdelta^u_t=\sqrt{\overline{\alpha_t}}\vx_0+\sqrt{1-\overline{\alpha_t}}\vepsilon_t+\sqrt{\overline{\alpha_t}}\vdelta^u_0
\end{equation}
Eq.~\ref{equ:perturb_decay} shows that as $t$ increases during the forward process, the impact of the protective noise will gradually decrease with decay coefficient $\sqrt{\overline{\alpha_t}}$. Since larger adversarial noise is more likely to have a greater impact on the model, the protective effect of the protective noise will decrease with the increase of $t$. 

\paragraph{Observation 2: Varying Importance of Timestep} Different step $t$ have different effects on the training process of DDPM since the scale of the noise added at each step $t$ during the forward process is different.
\begin{equation}
\label{equ:noise_scale_t}
    \lvert\nabla_t \vx_t\rvert = \lvert \vx_t - \vx_{t-1} \rvert = \left( \sqrt{\overline{\alpha_{t-1}}}-\sqrt{\overline{\alpha_t}} \right)\vx_0+\sqrt{\overline{\alpha_{t-1}}-\overline{\alpha_t}}\vepsilon_t
\end{equation}
Eq.~\ref{equ:noise_scale_t} illustrates the variation of the same image between two adjacent steps in the forward process, which implies that the scale of added-noise at step $t$ is $\sqrt{\overline{\alpha_{t-1}}-\overline{\alpha_t}}$. Following this, we can focus more on those $t$ with larger added noise when computing the protective noise considering that a step $t$ with a larger scale of added noise in the forward process may have a greater impact on the quality of the generated images. 

Given the above two observations, we propose Enhanced Unlearnable Diffusion Perturbation (EUDP), a method for computing protective noise with sampling timesteps based on the value of the production of the decay coefficient of protective noise $\sqrt{\overline{\alpha_t}}$ and the scale of added noise in the forward process $\sqrt{\overline{\alpha_{t-1}}-\overline{\alpha_t}}$. Specifically, when solving the max optimization problem for the parameter $\vdelta^u$ in Eq.~\ref{equ:UDP_maxmin}, the uniform distribution of $t$ is replaced with $\mathcal{P}_{\text{EUDP}}(t)$, a distribution based on the production $\sqrt{\overline{\alpha_t}}\cdot \sqrt{\overline{\alpha_{t-1}}-\overline{\alpha_t}}$:
\begin{equation}
\label{equ:EUDP_max}
    \begin{split}
        &\vdelta^u=\arg \max_{\Vert\vdelta^u_i\Vert\leq\rho_u} \left( \sum_{\vx_i} {\mathbb{E}_{\vx_{i,t_n}\sim p_\vtheta(\vx), t_n\sim \mathcal{P}_{\text{EUDP}}(t)} \mathcal{L}_{t_n}\left(f_\vtheta(\vx_{i,t_n}+\vdelta^u_i)\right)} \right) \\
        &\text{where} \quad \mathcal{P}_{\text{EUDP}}(t)=\frac{\sqrt{\overline{\alpha_t}}\cdot \sqrt{\overline{\alpha_{t-1}}-\overline{\alpha_t}}}{\sum_{t}{\left(\sqrt{\overline{\alpha_t}}\cdot \sqrt{\overline{\alpha_{t-1}}-\overline{\alpha_t}}\right)}} 
    \end{split}
\end{equation}
The sampling of timesteps in the minimization of $\vtheta$ in Eq.~\ref{equ:UDP_maxmin} is still following a uniform distribution. Protective noise $\vdelta^u$ and model parameter $\vtheta$ in the bi-level max-min optimization is solved by the iterative method in Eq.~\ref{equ:UDP_iter}. The pseudo-code of EUDP of DDPM is demonstrated in Appendix ~\ref{sec:Implementation of EUDP}.

\section{Experiments}

\begin{table}[htbp]
 \vspace{-2mm}
  \caption{Quality of images generated by DDPM trained on CIFAR-10 with different protective methods. Higher FID and lower Precision indicate lower quality and diversity. The percentages in parentheses demonstrate the increase in FID as well as the decrease in Precision and Recall of our methods compared with AdvDM.}
  \label{main exp}
  \centering
  \footnotesize
  \adjustbox{width=1\textwidth}{
  \begin{tabular}{ccccccc}
    \toprule
    \multirow{2}{*}{\textbf{Noise Scale}} & \multirow{2}{*}{\textbf{Metric}} & \multicolumn{3}{c}{\textbf{Methods}} &
    \multicolumn{2}{c}{\textbf{Methods (ours)}}
    \\
    \cmidrule(r){3-5} \cmidrule(r){6-7} & & Clean  & Random & AdvDM & UDP & EUDP \\
    \midrule
    \multirow{3}{*}{16/255} & FID$\uparrow$ & 3.83   & 31.50 & 34.01  & 56.52 (+66.19\%) & \textbf{60.75} (+78.62\%) \\
    & Precision(\%)$\downarrow$ & 71.77 & 45.12 & 47.60 & 36.04 (-24.29\%) & \textbf{30.88} (-35.13\%) \\
    & Recall(\%)$\downarrow$ & 54.13 & 40.25 & 36.93 & 29.58 (-19.90\%) & \textbf{23.81} (-35.53\%) \\
    \midrule
    \multirow{3}{*}{32/255} & FID$\uparrow$ & 3.83 & 65.13 & 79.91 & 106.71 (+33.54\%) & \textbf{112.67} (+41.00\%)\\
    & Precision(\%)$\downarrow$ & 71.77 & 25.25 & 35.94 & 31.27 (-13.00\%) & \textbf{23.67} (-34.14\%)\\
    & Recall(\%)$\downarrow$ & 54.12 & 22.30 & 20.09 & 12.37 (-38.43\%) & \textbf{10.47} (-47.88\%) \\
    \bottomrule
  \end{tabular}
  }
\end{table}

To evaluate the effectiveness of our method, we conduct a complete training (from scratch) on CIFAR-10~\citep{krizhevsky2009learning} dataset and utilize quantitative evaluations with metrics FID~\citep{heusel2017gans}, Precision, and Recall~\citep{sajjadi2018assessing,kynkaanniemi2019improved} to assess the quality of generated images.  We also demonstrate the efficacy of EUDP in fine-tuning the LDM in Appendix~\ref{sec:exp_ldm}. We compare the quality of images generated by DDPM trained on different CIFAR-10, including clean, random noise added, AdvDM~\citep{liang2023adversarial}, UDP, and EUDP. We evaluate each DDPM by generating 50,000 images and testing their FID, Precision, and Recall, where a smaller FID, larger Precision, and Recall indicate higher quality of generated images. 

Results demonstrated in Table~\ref{main exp} show that the FID of images generated by DDPM trained on the UDP dataset is significantly increased compared to random noise and AdvDM, while the Precision and Recall values are significantly reduced. This suggests that the UDP method effectively reduces the quality of generated images and successfully prevents the protected dataset from being used to train high-quality generative models. In addition, by using the EUDP method for protecting noise calculation, the quality of generated images is further reduced, indicating that EUDP can efficiently generate protective noise and achieve more effective results for protecting images.

\begin{wrapfigure}{r}{0.4\textwidth}
\vspace{-2mm}
  \includegraphics[width=0.4\textwidth]{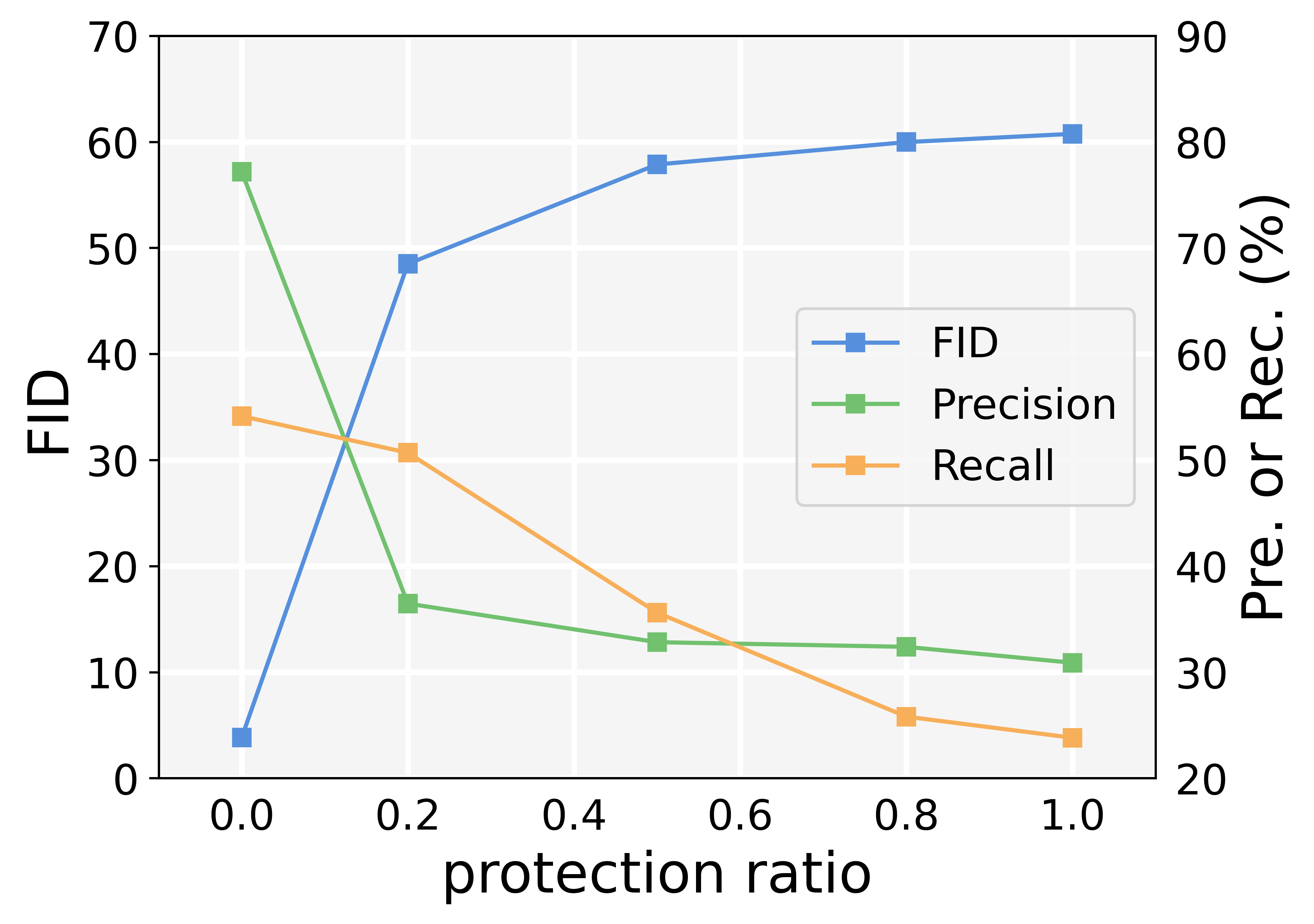}
  \vspace{-4mm}
  \caption{\label{fig:dataset poison} \footnotesize Quality of images generated by DDPM trained on EUDP CIFAR-10 with different protection ratio. FID increases while Precision and Recall decrease as the protection ratio increases.
  }
  \vspace{-4mm}
\end{wrapfigure}
We evaluate the protective effect at different protection ratios. Specifically, we randomly divided the CIFAR-10 into two sub-datasets in proportion: one clean and one protected. Images in the protected dataset are added with protected noise generated by EUDP. We combined the two sub-datasets to form the full training set with different protection ratios. The results in Figure~\ref{fig:dataset poison} show that the protective effect increases as the protection ratio increases, and the protective effect is significant when the protection ratio reached 50\%. We then add class-wise protective noise to the dataset. For the images of the 10 classes in the CIFAR-10, we only protect images with some of the classes. Table~\ref{class poison} shows the quality of the generated images of the corresponding protected and clean classes with class-wise added protective noise. The results show that the quality of the generated images of the protected classes is poor, while the quality of the generated images of the clean classes is similar to those generated on the clean CIFAR-10.

\begin{table}[htbp]
  \caption{Quality of images generated by DDPM trained on EUDP CIFAR-10 with different protected classes. The column of Protected Class indicates labels of protected samples in CIFAR-10 while Clean Class indicates labels of clean samples. Columns of Clean and Protected show results of DDPM trained on the clean CIFAR-10 and the class-wise protected CIFAR-10 respectively.}
  \label{class poison}
  \centering
  \footnotesize
  \adjustbox{width=0.9\textwidth}{
  \begin{tabular}{c|ccccccc}
    \toprule
    \textbf{Noise Scale} & \textbf{Metric} & \textbf{Protected Class} & \textbf{Clean} & \textbf{Protected} & \textbf{Clean Class} & \textbf{Clean} & \textbf{Protected}  \\
    \midrule
    \multirow{9}{*}{16/255} & FID$\uparrow$ & \multirow{3}{*}{$0\sim1$} & 8.55 & 58.17 & \multirow{3}{*}{$2\sim9$} & 4.27 & 7.81 \\
    & Precision(\%)$\downarrow$ & & 70.37 & 32.13 & & 70.09 & 67.45\\
    & Recall(\%)$\downarrow$ & & 53.34 & 28.70 & & 54.12 & 53.97\\
    \cmidrule(r){2-8}
    & FID$\uparrow$ & \multirow{3}{*}{$0\sim4$} & 5.48 & 47.16 & \multirow{3}{*}{$5\sim9$} & 5.44 & 8.04 \\
    & Precision(\%)$\downarrow$ & & 68.31 & 41.25 & & 72.61 & 68.60\\
    & Recall(\%)$\downarrow$ & & 54.70 & 32.06 & & 53.47 & 54.05\\
    \cmidrule(r){2-8}
    & FID$\uparrow$ & \multirow{3}{*}{$0\sim7$} & 4.48 & 41.48 & \multirow{3}{*}{$8\sim9$} & 6.31 & 8.42 \\
    & Precision(\%)$\downarrow$ & & 70.05 & 45.44 & & 71.08 & 66.97\\
    & Recall(\%)$\downarrow$ & & 54.27 & 32.21 & & 51.90 & 52.83\\
    \bottomrule
  \end{tabular}
  }
\end{table}

\section{Conclusion}
In this paper, we propose image-protection methods UDP and EUDP for diffusion image generative models from a perspective of unlearnable examples which protects the privacy and copyright of the image owners. We demonstrate that our method can successfully achieve the goal of making the protected images ``unlearnable'' and has practical applications.

\section*{Acknowledgement}
This work is partially supported by the National Key R\&D Program of China (under Grant 2021ZD0110102), the NSF of China (under Grants 62002338, 61925208, U22A2028, 62222214, 62102399, U19B2019), CAS Project for Young Scientists in Basic Research (YSBR-029), Youth Innovation Promotion Association CAS and Xplore Prize.

\bibliography{iclr2024_conference}
\bibliographystyle{iclr2024_conference}
\clearpage
\appendix
\section{Related Works}
\subsection{Diffusion Models}
Denoising Diffusion Probabilistic Model (DDPM)~\citep{sohl2015deep,ho2020denoising} is an image-generative diffusion model including a forward diffusion process and a reverse denoising process. In the forward process, an image $\vx_0$ is gradually perturbated with random Gaussian noise for $T$ steps and finally turns into random noise. In the reverse process, an image is conversely generated by gradually removing noise. Images at step $t$ in the forward process can be expressed by images at step $t-1$: $\vx_{t}=\sqrt{\alpha_{t}}\vx_{t-1}+\sqrt{\beta_{t}}\vepsilon$, where $\alpha_{t}$ and $\beta_{t}=1-\alpha_{t}$ are usually pre-defined parameters, and $\vepsilon$ following standard Gaussian distribution $\mathcal{N}(\vzero,\vI)$. Noise $\vepsilon_\vtheta$ denoised in the reverse process is learned by minimizing the following simplified loss function:
\begin{equation}
 \label{equ:dm_loss_simple}
\mathcal{L}_{\text{simple}}=\mathbb{E}_{t,\vx_0,\vepsilon}\left[{\Vert \vepsilon - \vepsilon_\vtheta(\sqrt{\bar{\alpha}_t} \vx_0 + \sqrt{1-\bar{\alpha}_t}\vepsilon, t) \Vert}^2 \right]
\end{equation}
Latent Diffusion Models (LDM)~\citep{rombach2022high} transfer diffusion processes from pixel space to latent space and introduce a cross-attention layer into model architecture to generate high-resolution images with general conditioning inputs. Latent noise denoised in the reverse process is learned by minimizing the following LDM loss function:
\begin{equation}
 \label{equ:ldm_loss}
\mathcal{L}_{\text{LDM}}=\mathbb{E}_{t,\vx_0,\vepsilon}\left[{\Vert \vepsilon - \vepsilon_\vtheta(\vz_t, t, \tau_\vtheta(\vy)) \Vert}^2 \right]
\end{equation}
Compared with Eq.~\ref{equ:dm_loss_simple}, an image $\vx$ is encoded into a latent vector by a pre-trained auto encoder $\mathcal{E}(\vx)$ and the output $\vz$ of the reverse process is decoded into an image by a decoder $\mathcal{D}(\vz)$. A conditioning input $y$ could be a set of text or an image and is encoded into a conditioning vector by a domain-specific encoder $\tau_\vtheta$. To achieve the unlearnable examples for diffusion models, the aim of our UDP and EUDP is to hinder the optimization of loss in Eq.~\ref{equ:dm_loss_simple} and ~\ref{equ:ldm_loss}.
\subsection{Unlearnable Examples for Classification Models}
Recent research~\citep{sandoval2022poisons,huang2021unlearnable,fu2022robust,yu2022availability,tao2021better} show that poison attack on neural networks based classification models can effectively decrease the accuracy of classifiers on the test set. Unlearnable examples are generated by adding imperceptible adversarial noise to a clean dataset and classifiers trained on unlearnable examples fail to generalize to the unseen samples. The main approaches to generate unlearnable examples include error-minimization and error-maximization.
\paragraph{Error-Minimization Noise}
The basic idea of error-minimization noise~\citep{huang2021unlearnable,fu2022robust} is to reduce the training loss of classifiers considering that less loss corresponds to less knowledge to learn.
\begin{equation}
 \label{equ:error_minimize}
\min_\vtheta \frac{1}{n} \sum^n_{i=1} \min_{\Vert \vdelta_i \Vert<\rho} {\mathcal{L}(f_\vtheta(\vx_i+\vdelta_i), y_i)}
\end{equation}
As shown in Eq.~\ref{equ:error_minimize}, the error-minimization noise is generated by solving a bi-level min-min optimization problem, where the inner minimization is to find the bounded protected noise that minimizes the classification loss, while the outer minimization finds model parameters that also minimize the loss of the classifier.
\paragraph{Error-Maximization Noise}
~\citep{fowl2021adversarial} shows that error-maximization noise generated during adversarial training in Eq.~\ref{equ:error_maximize} is also significant to poor the performance of classifiers. The generation process of error-maximization noise is the same as adversarial training, which solves a bi-level min-max optimization problem. Different from the error-minimization noise, inner maximization finds noise that maximizes the loss of the classifier.
\begin{equation}
 \label{equ:error_maximize}
\max_\vtheta \frac{1}{n} \sum^n_{i=1} \min_{\Vert \vdelta_i \Vert<\rho} {\mathcal{L}(f_\vtheta(\vx_i+\vdelta_i), y_i)}
\end{equation}
However, there is no general definition of unlearnable examples for generative models, which may hold greater practical significance compared to classification models due to the privacy concerns brought by AIGC.

\subsection{Security \& Privacy Protection for Diffusion Models}
\paragraph{Adversarial Attacks for Diffusion Models} Adversarial attacks~\citep{goodfellow2014explaining,xu2018structured,xu2020adversarial,zhang2022branch,li2019nattack} have been introduced to diffusion models recently: \citep{salman2023raising} presents a targeted adversarial attack algorithm towards Latent Diffusion Models including encoder attack and diffusion attack. The encoder attack is to generate adversarial perturbations $\vdelta$ by optimizing the distance between the latent code of perturbated samples $\mathcal{E}(\vx+\vdelta)$ and a target latent vector $z_\text{targ}$ while the diffusion attack optimizes the distance between samples generated by LDM and target images. This method can effectively prevent images from being modified by diffusion models in an image-to-image fashion. Liang et al. propose AdvDM~\citep{liang2023adversarial}, an untargeted adversarial examples generation method for diffusion models through Monte Carlo. AdvDM randomly samples different timesteps and latent variables to iteratively upgrade the adversarial noise. Though adversarial examples generated by AdvDM can partially disrupt the training of diffusion models, incorporating the training process of diffusion models into the generation of protective noise can greatly enhance this disruption.
\paragraph{Privacy Protection for Generative Models} There is an increasing demand for privacy protection for generative models because of the fast development of image synthesis methods. ~\citep{gandikota2023erasing} achieves the goal of preventing Diffusion Models from generating privacy-sensitive images by erasing the concepts within the models. Recent detection methods for DeepFakes~\citep{fernandez2023stable,guo2023hierarchical,jiang2023evading,wang2023dire,sha2022fake,abdelnabi2021adversarial,frank2020leveraging,zhao2021multi} can also help prevent the abuse of generative models, especially Diffusion Models, by distinguishing whether an image is generated by them. A more direct approach is to prevent images from being used to train image generative models. ~\citep{shan2023glaze} proposes Glaze, a targeted adversarial attack on the feature extractor of text-to-image models, protecting artworks from style mimicry by misleading the match between text feature vectors and image feature vectors. In comparison to our method, this work does not specifically target the diffusion process, but focuses on text-to-image generative models and specifically addresses style transfer tasks. ~\cite{van2023anti} propose Anti-DreamBooth to protect personal face images from being used to train LDM via the fine-tuning method DreamBooth. This work focuses on the specific fine-tuning scenario and the LDM but does not discuss it from an unlearnable aspect and the diffusion process itself.
\section{Threat Model}
We consider two parties involved in the process: the Image Exploiter for diffusion model training and the Image Protector for the IP owner. We aim to design Image Protector to prevent the selected images from being utilized to train or fine-tune high-quality image models but with no harm to the available public data from being utilized. Specifically, we explain the workflow of Image Exploiter and Image Protector as follows:

\paragraph{Image Exploiter} The Image Exploiter creates a training dataset for image generative models training or finetuning based on gathered public image resources. In practical settings, fine-tuning Diffusion Models usually includes two types: training specific objects or training specific styles. The former involves fine-tuning the model using a few images of a specific object to generate various images related to that object, while the latter involves training the model using a few images of a specific style for tasks such as style transfer. Generally, fine-tuning LDM requires the following knowledge: (1) a pre-trained LDM model, including the model structure and all parameters; (2) images of specific objects or styles, while IP of some styles or objects may be owned by other parties and unauthorized data exploitation should be eliminated.
\paragraph{Image Protector} The Image Protector aims to prevent their images from being used to train or fine-tune high-quality image generative models (such as LDM) while still making them publicly available. Specifically, the Image Protector wishes to prevent their images from being used to generate false images of specific objects (DeepFake) or images of specific styles (copyright infringement). We assume that the Image Protector possesses the following knowledge: (1) the images that need to be protected; (2) the image generative models that the Image Exploiter may use, including the models' structures and parameters. This assumption is practically significant due to the fact that many current pre-trained LDMs are fine-tuned on widely used LDMs such as \textit{stable-diffusion-v1-4}. Additionally, we conducted transferability experiments on LDMs to demonstrate that this assumption can be relaxed.
\section{Implementation Details}
\label{sec:Implementation Details}
\subsection{Implementation of UDP}
\label{sec:Implementation of UDP}
We demonstrate the pseudo-code of Unlearnable Diffusion Perturbation for DDPM and LDM in Algorithm~\ref{alg:UDP_DDPM} and Algorithm~\ref{alg:UDP_LDM} respectively. As shown in Eq.~\ref{equ:UDP_maxmin} and Eq.~\ref{equ:UDP_iter}, the bi-level max-min optimization problem can be solved by alternatively optimizing the parameters $\vtheta$ of the model and the protective noise $\vdelta^u$. Specifically, the parameters $\vtheta$ are firstly optimized by minimizing the loss for $K$ steps with learning rate $\eta$ and then the perturbations $\vdelta^u$ are iteratively calculated by maximizing the loss for $M$ steps with perturbation rate $\lambda$. After each step of perturbation calculation, $\vdelta^u$ will be clipped with the noise scale $\rho_u$. This bi-level optimization process will be iterated for $N$ times until the perturbation can effectively protect images. Here $K$, $M$, $N$, $\eta$, and $\rho_u$ are pre-defined hyper-parameters and the perturbation rate $\lambda$ is typically set to 1/10 of the noise scale $\rho_u$. The implementations of UDP for DDPM and UDP for LDM are similar, with the main difference being that the loss in LDM (Eq.~\ref{equ:dm_loss_simple}) is computed in the latent space and includes conditioning guidance $\vy$ compared with DDPM (Eq.~\ref{equ:ldm_loss}).

\begin{CJK}{UTF8}{gbsn} 
\begin{algorithm} [htbp]
\caption{Unlearnable Diffusion Perturbation for DDPM}
\label{alg:UDP_DDPM}
    \floatname{algorithm}{Algorithm}
    \begin{algorithmic}[1]
    \Require Clean Dataset $\vx_i\in \mathcal{D}_c$, U-Net Model $f_\vtheta$, Max Perturbation Scale $\rho_u$, Iteration Steps $N$, Learning Rate $\eta$, Training Steps $K$, Perturbation Rate $\lambda$, Perturbation Steps $M$
    \Ensure Protective Noise $\vdelta^u$
        \State $\vtheta \leftarrow \vtheta_0$
        \State $\vdelta^u \leftarrow \vdelta^u_0$
        \For{$n = 1 \rightarrow N$}
            \For{$k = 1 \rightarrow K$}
                \State \textbf{sample} $t \sim \mathcal{U}\left(0,T\right)$, $\vx_i \in \mathcal{D}_c$
                \State $\vtheta \leftarrow \vtheta-\eta \cdot \nabla_\vx \mathcal{L}_t\left( f_\vtheta(\vx_{i,t}+\vdelta^u_i) \right) $
            \EndFor
            \For{$m = 1 \rightarrow M$}
                \For{$\vx_i$ in $\mathcal{D}_c$}
                    \State \textbf{sample} $t \sim \mathcal{U}\left(0,T\right)$
                    \State $\vdelta^u_i \leftarrow \vdelta^u_i+\lambda \cdot \text{sign} \left( \nabla_\vx \mathcal{L}_{t_n}\left( f_\vtheta(\vx_{i,t_n}+\vdelta^u_i) \right)\right) $
                    \If {$\Vert\vdelta^u_i\Vert > \rho_u$}
                        \State $\vdelta^u_i \leftarrow \text{sign}\left(\vdelta^u_i\right)\cdot\rho_u$
                    \EndIf
                \EndFor
            \EndFor
        \EndFor
        \State \Return $\vdelta^u_i$
    \end{algorithmic} 
\end{algorithm}
\end{CJK} 

\begin{CJK}{UTF8}{gbsn} 
\begin{algorithm} [htbp]
\caption{Unlearnable Diffusion Perturbation for LDM}
    \label{alg:UDP_LDM}
    \floatname{algorithm}{Algorithm}
    \begin{algorithmic}[1]
    \Require Clean Dataset $\vx_i\in \mathcal{D}_c$, pre-trained U-Net Model $f_\vtheta$, pre-trained Encoder $\mathcal{E}$, Text Encoder $\tau_\vtheta$, Conditional Guidance $\vy$, Max Perturbation Scale $\rho_u$, Iteration Steps $N$, Learning Rate $\eta$, Training Steps $K$, Perturbation Rate $\lambda$, Perturbation Steps $M$
    \Ensure Protective Noise $\vdelta^u$
        \State $\vtheta \leftarrow \vtheta_0$
        \State $\vdelta^u \leftarrow \vdelta^u_0$
        \For{$n = 1 \rightarrow N$}
            \For{$k = 1 \rightarrow K$}
                \State \textbf{sample} $t \sim \mathcal{U}\left(0,T\right)$, $\vx_i \in \mathcal{D}_c$
                \State $\vtheta \leftarrow \vtheta-\eta \cdot \nabla_\vx \mathcal{L}_t\left( f_\vtheta\left(\mathcal{E}\left(\vx_{i,t}+\vdelta^u_i\right), \tau_\vtheta\left(\vy\right)\right) \right) $
            \EndFor
            \For{$m = 1 \rightarrow M$}
                \For{$\vx_i$ in $\mathcal{D}_c$}
                    \State \textbf{sample} $t \sim \mathcal{U}\left(0,T\right)$
                    \State $\vdelta^u_i \leftarrow \vdelta^u_i+\lambda \cdot \text{sign} \left( \nabla_\vx \mathcal{L}_{t_n}\left( f_\vtheta\left(\mathcal{E}\left(\vx_{i,t}+\vdelta^u_i\right), \tau_\vtheta\left(\vy\right)\right) \right)\right) $
                    \If {$\Vert\vdelta^u_i\Vert > \rho_u$}
                        \State $\vdelta^u_i \leftarrow \text{sign}\left(\vdelta^u_i\right)\cdot\rho_u$
                    \EndIf
                \EndFor
            \EndFor
        \EndFor
        \State \Return $\vdelta^u_i$
    \end{algorithmic} 
\end{algorithm}
\end{CJK}

\subsection{Implementation of EUDP}
\label{sec:Implementation of EUDP}
The pseudo-code of Enhanced Unlearnable Diffusion Perturbation for DDPM and LDM is described in Algorithm~\ref{alg:EUDP_DDPM} and Algorithm~\ref{alg:EUDP_LDM} respectively. The implementation of EUDP is similar to UDP, but the sampling of timesteps $t$ during the optimization of protective noise $\vdelta^u$ follows the distribution of $\mathcal{P}_\text{EUDP}\left(t\right)$ as shown in Eq.~\ref{equ:EUDP_max} instead of the uniform distribution $\mathcal{U}(0,T)$.
\begin{CJK}{UTF8}{gbsn} 
\begin{algorithm} [htbp]
\caption{Enhanced Unlearnable Diffusion Perturbation for DDPM}
    \label{alg:EUDP_DDPM}
    \floatname{algorithm}{Algorithm}
    \begin{algorithmic}[1]
    \Require Clean Dataset $\vx_i\in \mathcal{D}_c$, U-Net Model $f_\vtheta$, Max Perturbation Scale $\rho_u$, Iteration Steps $N$, Learning Rate $\eta$, Training Steps $K$, Perturbation Rate $\lambda$, Perturbation Steps $M$
    \Ensure Protective Noise $\vdelta^u$
        \State $\vtheta \leftarrow \vtheta_0$
        \State $\vdelta^u \leftarrow \vdelta^u_0$
        \For{$n = 1 \rightarrow N$}
            \For{$k = 1 \rightarrow K$}
                \State \textbf{sample} $t \sim \mathcal{U}\left(0,T\right)$, $\vx_i \in \mathcal{D}_c$
                \State $\vtheta \leftarrow \vtheta-\eta \cdot \nabla_\vx \mathcal{L}_t\left( f_\vtheta(\vx_{i,t}+\vdelta^u_i) \right) $
            \EndFor
            \For{$m = 1 \rightarrow M$}
                \For{$\vx_i$ in $\mathcal{D}_c$}
                    \State \textbf{sample} $t \sim \mathcal{P}_\text{EUDP}\left(t\right)$
                    \State $\vdelta^u_i \leftarrow \vdelta^u_i+\lambda \cdot \text{sign} \left( \nabla_\vx \mathcal{L}_{t_n}\left( f_\vtheta(\vx_{i,t_n}+\vdelta^u_i) \right)\right) $
                    \If {$\Vert\vdelta^u_i\Vert > \rho_u$}
                        \State $\vdelta^u_i \leftarrow \text{sign}\left(\vdelta^u_i\right)\cdot\rho_u$
                    \EndIf
                \EndFor
            \EndFor
        \EndFor
        \State \Return $\vdelta^u_i$
    \end{algorithmic} 
\end{algorithm}
\end{CJK} 

\begin{CJK}{UTF8}{gbsn} 
\begin{algorithm} [htbp]
\caption{Enhanced Unlearnable Diffusion Perturbation for LDM}
    \label{alg:EUDP_LDM}
    \floatname{algorithm}{Algorithm}
    \begin{algorithmic}[1]
    \Require Clean Dataset $\vx_i\in \mathcal{D}_c$, pre-trained U-Net Model $f_\vtheta$, pre-trained Encoder $\mathcal{E}$, Text Encoder $\tau_\vtheta$, Condition Guidance $\vy$, Max Perturbation Scale $\rho_u$, Iteration Steps $N$, Learning Rate $\eta$, Training Steps $K$, Perturbation Rate $\lambda$, Perturbation Steps $M$
    \Ensure Protective Noise $\vdelta^u$
        \State $\vtheta \leftarrow \vtheta_0$
        \State $\vdelta^u \leftarrow \vdelta^u_0$
        \For{$n = 1 \rightarrow N$}
            \For{$k = 1 \rightarrow K$}
                \State \textbf{sample} $t \sim \mathcal{U}\left(0,T\right)$, $\vx_i \in \mathcal{D}_c$
                \State $\vtheta \leftarrow \vtheta-\eta \cdot \nabla_\vx \mathcal{L}_t\left( f_\vtheta\left(\mathcal{E}\left(\vx_{i,t}+\vdelta^u_i\right), \tau_\vtheta\left(\vy\right)\right) \right) $
            \EndFor
            \For{$m = 1 \rightarrow M$}
                \For{$\vx_i$ in $\mathcal{D}_c$}
                    \State \textbf{sample} $t \sim \mathcal{P}_\text{EUDP}\left(t\right)$
                    \State $\vdelta^u_i \leftarrow \vdelta^u_i+\lambda \cdot \text{sign} \left( \nabla_\vx \mathcal{L}_{t_n}\left( f_\vtheta\left(\mathcal{E}\left(\vx_{i,t}+\vdelta^u_i\right), \tau_\vtheta\left(\vy\right)\right) \right)\right) $
                    \If {$\Vert\vdelta^u_i\Vert > \rho_u$}
                        \State $\vdelta^u_i \leftarrow \text{sign}\left(\vdelta^u_i\right)\cdot\rho_u$
                    \EndIf
                \EndFor
            \EndFor
        \EndFor
        \State \Return $\vdelta^u_i$
    \end{algorithmic} 
\end{algorithm}
\end{CJK} 
\section{Experiment Settings}
\label{sec:Experiment Settings}
\subsection{Parameters of Diffusion Models}
\label{sec:Parameters of Diffusion Models}
We train a DDPM from scratch and fine-tune an LDM with a pre-trained model. For DDPM training on CIFAR-10, we set the batch size to $128$, the learning rate to $0.0001$, and the number of epochs to $2000$ ($\sim 80k$ steps in total). The noise scheduler in the diffusion process is the widely used linear scheduler with $\beta_0=0.0001$ and $\beta_T=0.02$ where $T=1000$. For LDM, we utilize \textit{stable-diffusion-v1-4} \footnote{\url{https://huggingface.co/CompVis/stable-diffusion-v1-4}} as the base model and fine-tune it with different parameters for DreamBooth~\citep{ruiz2022dreambooth} and Textual Inversion~\citep{gal2022image}. Specifically, we fine-tune the model with DreamBooth with batch size set to $1$, learning rate set to $4\times10^{-6}$, and steps to $400$ and Textual Inversion with batch size set to $1$, learning rate set to $5\times10^{-4}$, and steps to $5000$.
\subsection{Parameters of UDP \& EUDP}
\label{sec:Parameters of UDP and EUDP}
For UDP and EUDP for DDPM in Algorithm~\ref{alg:UDP_DDPM} and~\ref{alg:EUDP_DDPM}, we set parameters $N$ to $100$, $K$ to $1000$, $M$ to $10$, and $\eta$ to $0.0001$. Due to the significant training cost, we do not conduct systematic ablation experiments on these parameters. Empirically, we find that setting $N\times M\times \lambda$ to 100 times of the scale of the noise $\rho_u$ can ensure the protective effect of the perturbation. For experiments of LDM, we set $N$ to $40$, $K$ to $20$, $M$ to $25$, and $\eta$ to $5\times10^{-6}$ in Algorithm~\ref{alg:UDP_LDM} and ~\ref{alg:EUDP_LDM}. The noise scale of protective noise in all experiments of LDM is 16/255.
\subsection{Other Settings}
\label{sec:Other Settings}
We use the \textit{diffuser} library \footnote{\url{https://github.com/huggingface/diffusers}} for the training and inference of diffusion models and utilize this code \footnote{\url{https://github.com/openai/guided-diffusion/tree/main/evaluations}} to evaluate the quality of generated images with FID, Precision, and Recall. For the baseline AdvDM~\citep{liang2023adversarial} in the quantitative experiments of DDPM, we set the iteration steps to $128$ to ensure sufficient generation of adversarial noise.
\section{Results on Latent Diffusion Model}
\label{sec:exp_ldm}
\begin{figure}[htbp]
  \centering
  \subfigure[Clean and protected images for fine-tuning a Stable Diffusion model]{
    \includegraphics[width=1\textwidth]{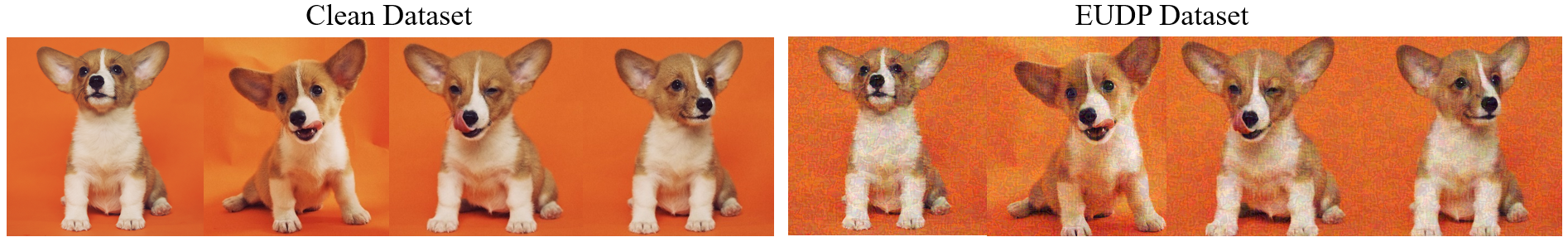}
    \label{fig:dataset_text2img}
  }
  \subfigure[Generated images of Stable Diffusion fine-tuned with DreamBooth]{
    \includegraphics[width=0.47\textwidth]{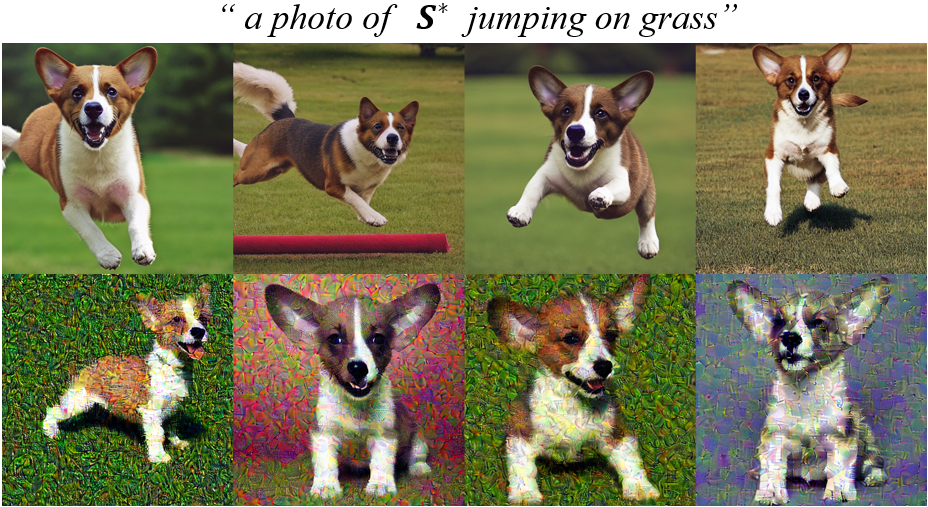}
    \label{fig:db_text2img}
  }
  \quad
  \subfigure[Generated images of Stable Diffusion fine-tuned with Textual Inversion]{
    \includegraphics[width=0.47\textwidth]{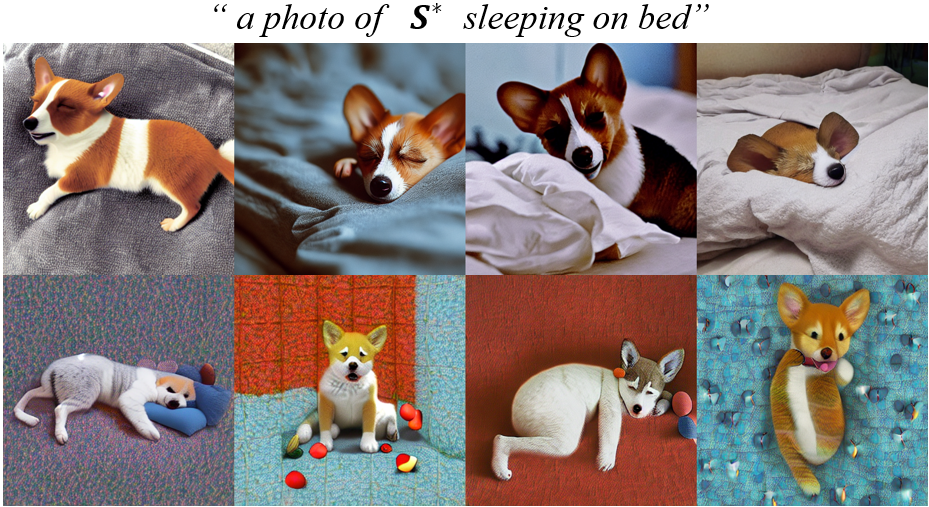}
    \label{fig:ti_text2img}
  }
  \vspace{-3mm}
  \caption{\label{fig:text2object} An example of text-to-image with object protection. (b)\&(c): The first row: Generated images of Stable Diffusion fine-tuned on the clean dataset. The second row: Generated images of Stable Diffusion fine-tuned on the EUDP dataset.}
\end{figure}

\begin{figure}[htbp]
  \centering
  \includegraphics[width=1\textwidth]{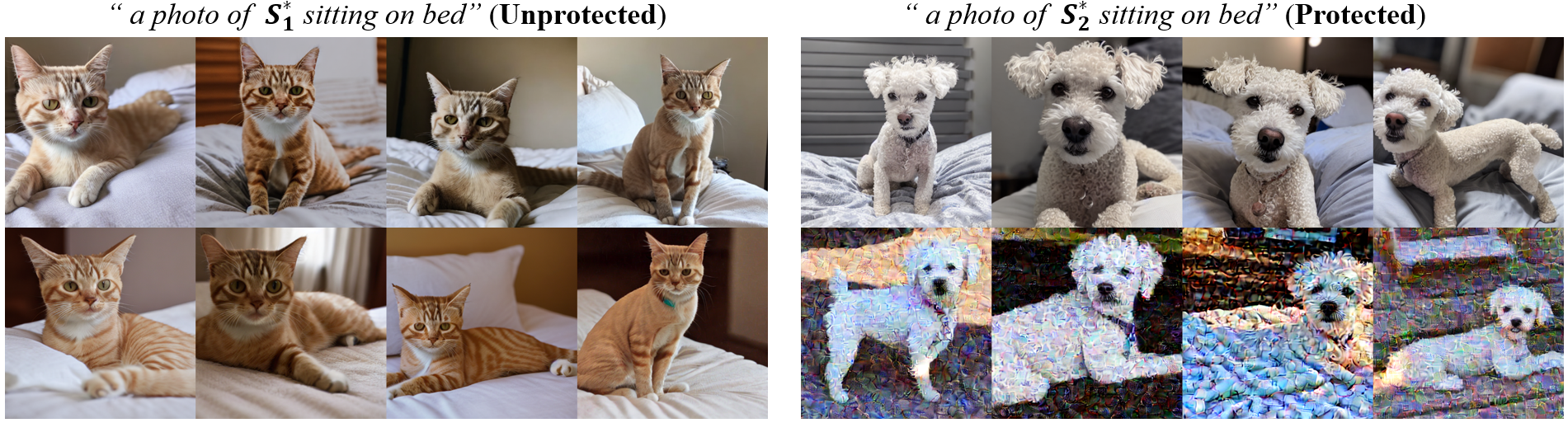}
  \caption{\label{fig:text2object_class} An example of text-to-image with class-wise protection. The first row: Generated images of Stable Diffusion fine-tuned on the clean dataset. The second row: Generated images of Stable Diffusion fine-tuned on class-wise EUDP dataset, where training images with label $\bm{S^*_1}$ are clean while training images with label $\bm{S^*_2}$ are protected by EUDP.}
\end{figure}

For LDM experiments, we fine-tune a widely used pre-trained LDM \textit{stable-diffusion-v1-4}~\citep{rombach2022high} with datasets provided by DreamBooth~\citep{ruiz2022dreambooth} and WikiArt~\citep{wikiart2016wikiart}. In LDM experiments, we qualitatively demonstrate the effectiveness of our method through image visualization results. For the task of training an object, we fine-tune the LDM using the DreamBooth dataset with fine-tuning methods Textual Inversion and DreamBooth. As shown in Figure~\ref{fig:text2object}, the LDM fine-tuned on the clean dataset is able to generate images that match the given prompt. However, the LDM fine-tuned on the EUDP dataset failed to generate the expected images. Notably, images generated by Textual Inversion have nearly no features of the training images, whereas those generated by DreamBooth are of low quality and repeat the training set, failing to meet the prompt. This demonstrates that our method successfully protects images of specific objects from being used to train a high-quality LDM. Moreover, Figure~\ref{fig:text2object_class} shows that protecting a specific label hardly affects the generative quality of label fine-tuning on the clean dataset. 

\begin{figure}[htbp]
  \centering
  \includegraphics[width=1\textwidth]{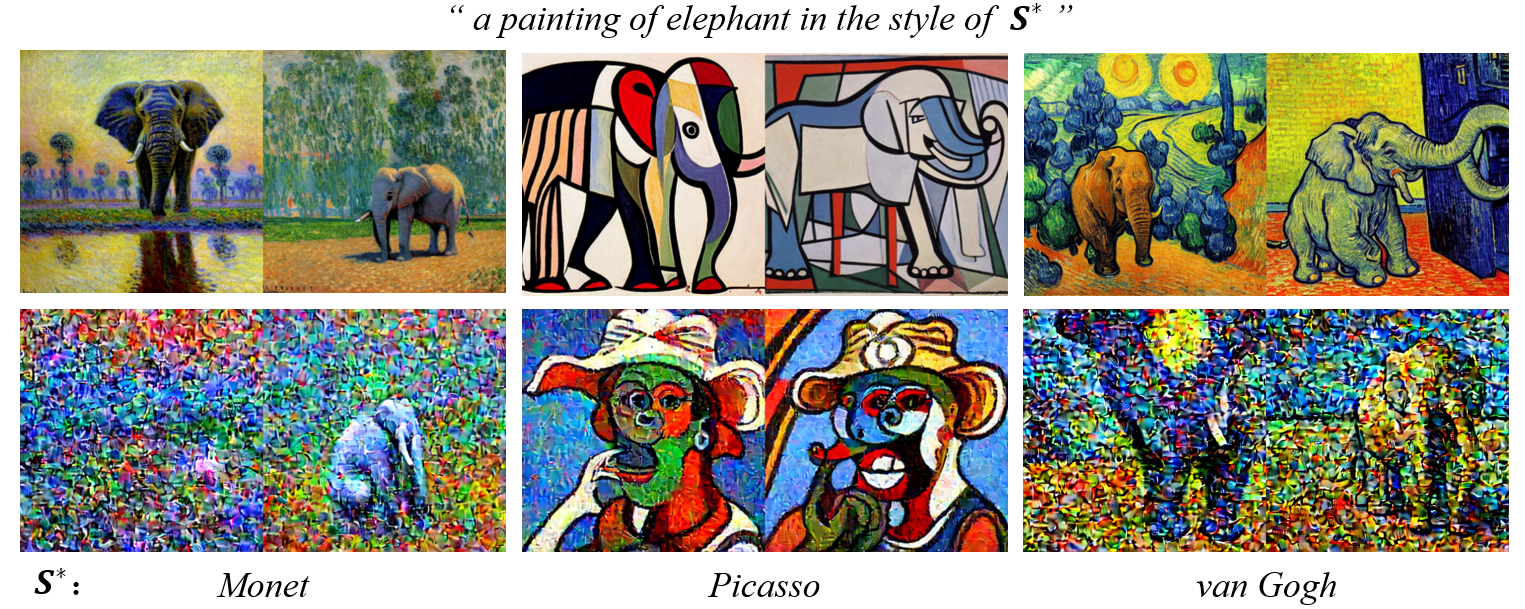}
  \vspace{-6mm}
  \caption{\label{fig:text2style} An example of text-to-image with specific style. The first row: Generated images of Stable Diffusion fine-tuned on the clean dataset. The second row: Generated images of Stable Diffusion fine-tuned on  the EUDP dataset.}
\end{figure}

\begin{figure}[htbp]
  \centering
  \includegraphics[width=1\textwidth]{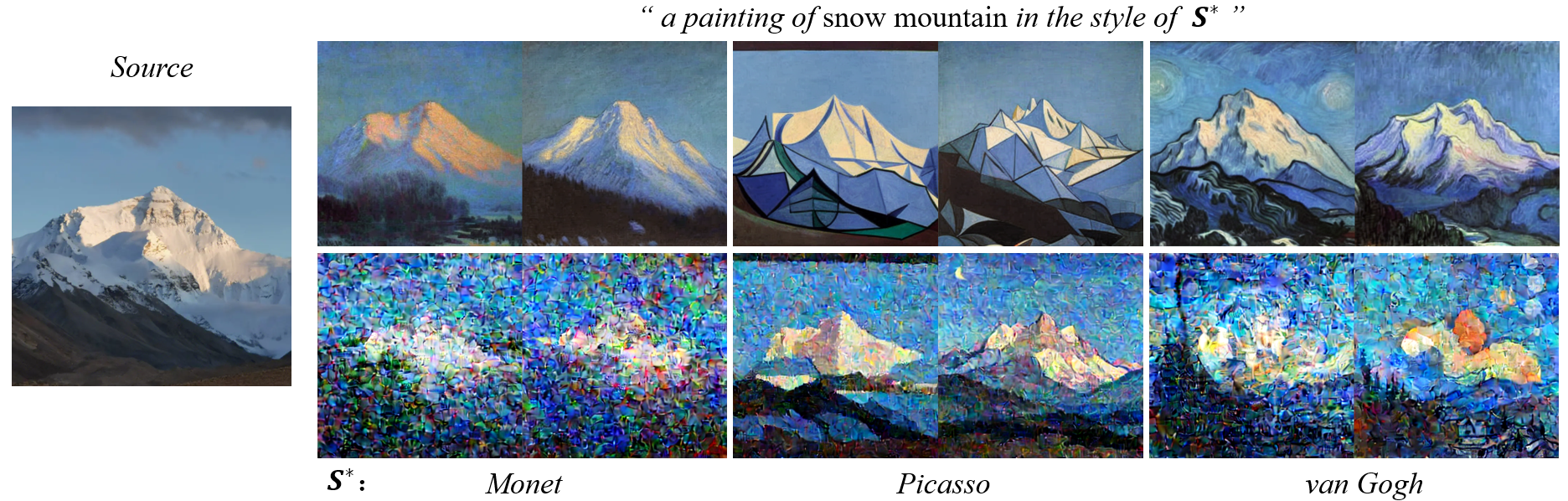}
   \vspace{-6mm}
  \caption{\label{fig:img2style} An example of style transfer. The first row: Generated images of Stable Diffusion fine-tuned on the clean dataset. The second row: Generated images of Stable Diffusion fine-tuned on the EUDP dataset.}
\end{figure}

In the scenario of training a style, we apply DreamBooth to fine-tune the LDM using the WikiArt dataset, evaluating it on both text-to-image and image-to-image (style transfer) tasks. Specifically, we selected six paintings by a particular artist (such as Monet) for style training. As shown in Figure~\ref{fig:text2style} and Figure~\ref{fig:img2style}, LDMs trained on the clean dataset are able to generate images with a specific artist's style and perform style transfer. Conversely, LDMs trained on the unlearnable dataset are unable to generate images with the corresponding style or convert the source image to a specific style. This experiment demonstrates that our method successfully protects specific styles, preventing infringement of copyright by unauthorized data exploitation such as style mimicry.

\section{Further Analysis}
\subsection{Robustness Study}
\label{sec:Robustness Study}
We evaluate the effectiveness of the protective noise after adding natural perturbations, including random noise, quantification, JPEG, and Gaussian blur. Specifically, the random noise has a scale of 16/255. Quantization involves reducing an 8-bit image to a 6-bit image. Gaussian blur is performed using a filter kernel with kernel size 4x4 and with $\sigma=$16/255. JPEG compression and decompression are carried out using the "imencode" and "imdecode" functions from the \textit{OpenCV2} library \footnote{\url{https://github.com/abidrahmank/OpenCV2-Python-Tutorials}}. Results in Table~\ref{robustness} show that JPEG compression and Gaussian blurring can improve the quality of generated images to some extent, but it is challenging to achieve the same level of high-quality generated images as in the (original) clean dataset.
\begin{table}[htbp]
  \caption{Quality of images generated by DDPM trained on CIFAR-10 with or without natural perturbation. }
  \label{robustness}
  \centering
  \small
  \begin{tabular}{ccccccc}
    \toprule
     \multirow{2}{*}{\textbf{Metric}} & \multicolumn{2}{c}{\textbf{Control}} & \multicolumn{4}{c}{\textbf{Natural Perturbation}}   \\
    \cmidrule(r){2-3} \cmidrule(r){4-7} & Clean & EUDP & Random Noise & JPEG & Quantify & Gaussian blur \\
    \midrule
    FID$\uparrow$ & 3.83 & 60.75 & 102.95 & 27.61 & 62.46 & 41.88 \\
    Precision(\%)$\downarrow$ & 71.77  & 30.88 & 20.78 & 55.23 & 31.35 & 60.59 \\
    Recall(\%)$\downarrow$ & 54.13 & 23.81 & 11.35 & 40.87 & 26.66 & 33.78 \\
    \bottomrule
  \end{tabular}
\end{table}
\subsection{Transferability Study}
\label{sec:Transferability Study}
\begin{figure}[htbp]
  \centering
  \subfigure[Generated images with \textit{stable-diffusion-v-1-4} trained on clean data]{
    \includegraphics[width=1\textwidth]{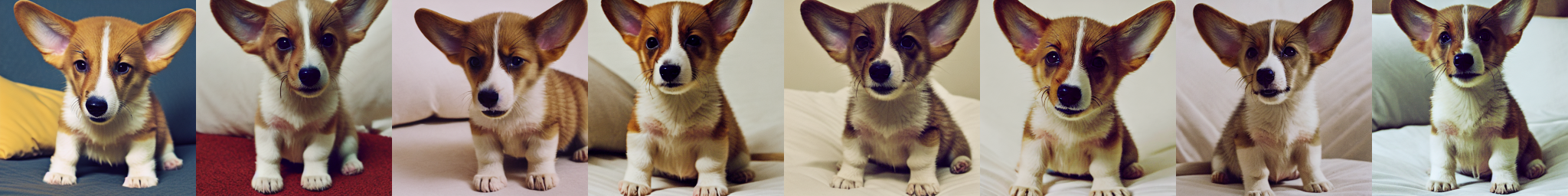}
    \label{fig:sd_v14}
  }
  \subfigure[Generated images with \textit{stable-diffusion-v-1-4} trained on protected data]{
    \includegraphics[width=1\textwidth]{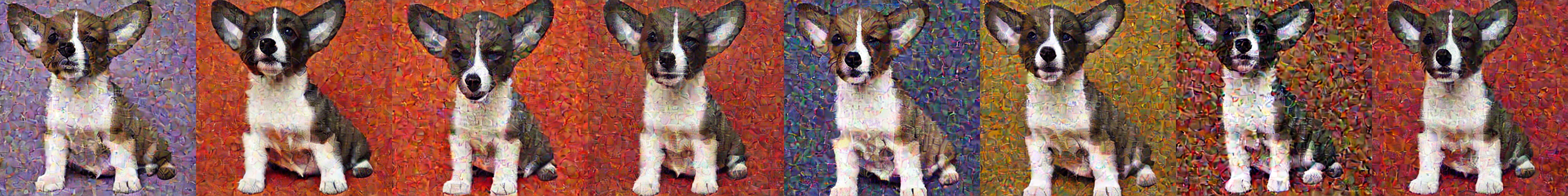}
    \label{fig:sd_v14_poisoned}
  }
  \subfigure[Generated images with \textit{stable-diffusion-v-1-1} trained on protected data]{
    \includegraphics[width=1\textwidth]{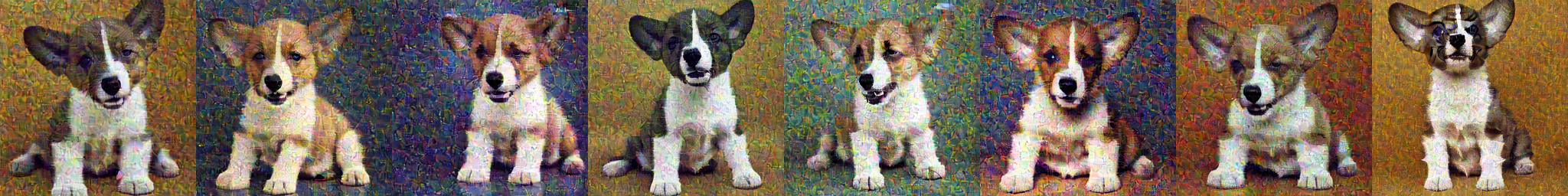}
    \label{fig:sd_v11_poisoned}
  }
  \subfigure[Generated images with \textit{stable-diffusion-v-1-5} trained on protected data]{
    \includegraphics[width=1\textwidth]{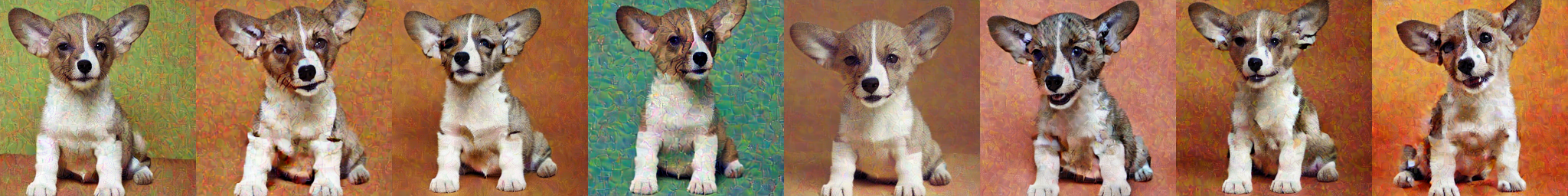}
    \label{fig:sd_v15_poisoned}
  }
  \subfigure[Generated images with \textit{Counterfeit-V2.5} trained on protected data]{
    \includegraphics[width=1\textwidth]{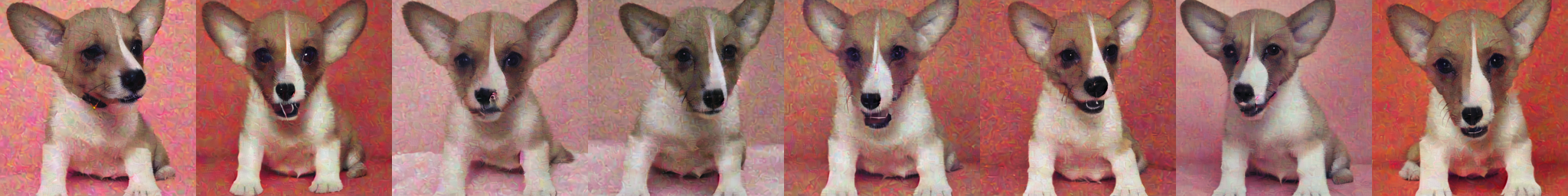}
    \label{fig:cf_v25_poisoned}
  }
  \vspace{-3mm}
  \caption{\label{fig:model_transferability} Generated images with different LDMs trained on clean or protected data and text promp for generation is \textit{"A photo of \textbf{$S^*$} dog sitting on bed"}. The protective noise of protected data is generated by EUDP with \textit{stable-diffusion-v1-4}. }
\end{figure}

\begin{figure}[htbp]
  \centering
  \subfigure[$S^* = $ \textit{AAA} ]{
    \includegraphics[width=1\textwidth]{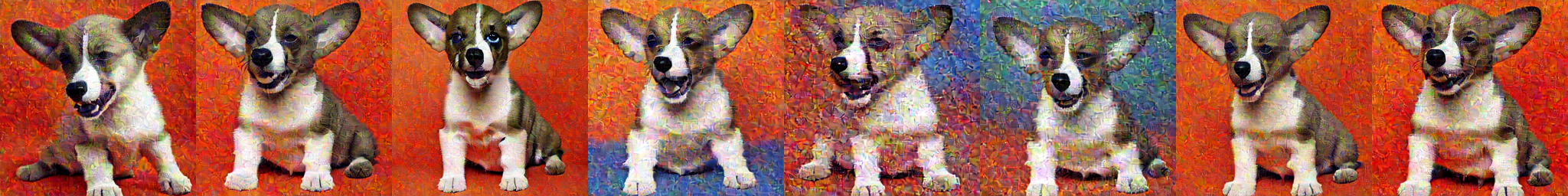}
    \label{fig:AAA}
  }
  \subfigure[$S^* = $ \textit{BBB}]{
    \includegraphics[width=1\textwidth]{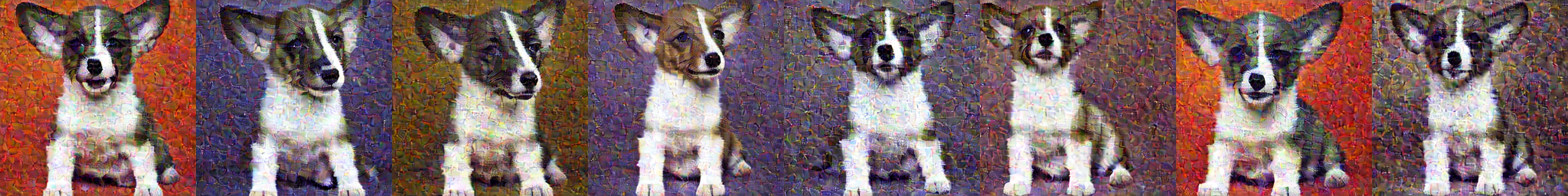}
    \label{fig:BBB}
  }
  \subfigure[$S^* = $ \textit{CCC}]{
    \includegraphics[width=1\textwidth]{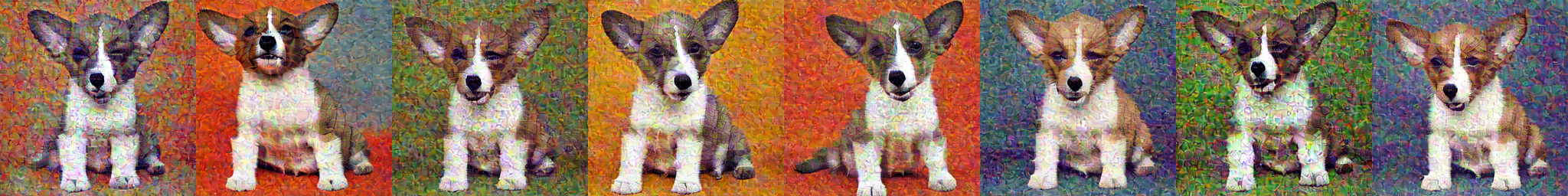}
    \label{fig:CCC}
  }
  \vspace{-3mm}
  \caption{\label{fig:prompt_transferability} Generated images with \textit{stable-diffusion-v1-4} trained on protected data. The text prompt for training is \textit{"A photo of $S^*$ dog"} and the text prompt for for generation is \textit{"A photo of $S^*$ dog sitting on bed"} while the text prompt for protective nosie generation is \textit{"A photo of sks dog"}.}
\end{figure}

We conduct transferability studies to evaluate the effectiveness of our methods in black-box conditions. We first assess the transferability of the noise scheduler of EUDP for DDPM. Protective noise is generated by EUDP with $\beta_0=0.0001$ and $\beta_T=0.02$ and the protected images are tested with other noise schedulers. Results demonstrated in Table~\ref{noise scheduler} indicate that changing the noise scheduler does not significantly diminish the effectiveness of the protection noise. Additionally, it can be observed from the experimental results that the EUDP method, corresponding to a noise scheduler with faster noise adding (larger $\beta_T$) during the training process, exhibits a greater improvement in protection compared to the UDP method, which also confirms the validity of our observations in Section~\ref{sec:EUDP}.
For real-world conditions, we conduct transferability studies of our methods for LDM. We first examine the transferability of EUDP between different LDMs. Specifically, the protective noise is generated with \textit{stable-diffusion-v1-4} and the protected images are learned by pre-trained LDMs \textit{stable-diffusion-v1-1} \footnote{\url{https://huggingface.co/CompVis/stable-diffusion-v1-1}}, \textit{stable-diffusion-v1-5}\footnote{\url{https://huggingface.co/runwayml/stable-diffusion-v1-5}}, and \textit{Counterfeit-V2.5}\footnote{\url{https://huggingface.co/gsdf/Counterfeit-V2.5}}. Results in Figure~\ref{fig:model_transferability} show that protective noise generated by a specific model remains effective in protecting images from being learned by other models. Furthermore, when the model used for training is closer to the model for protection noise generation, the protective effect is better (\textit{stable-diffusion-v1-1}). Conversely, when there is a significant difference between the two models, the protection effect is weaker (\textit{Counterfeit-V2.5}). Figure~\ref{fig:prompt_transferability} demonstrates that the text prompt used for protective noise generation can be different from the text prompt for fine-tuning diffusion models since our methods mainly focus on the diffusion process instead of the text encoder.

\begin{table}[htbp]
  \caption{Quality of images generated by DDPM trained on UDP or EUDP CIFAR-10 with different noise schedulers. The protective noise is generated with a noise scheduler where $\beta_0=0.0001$ and $\beta_T=0.02$.}
  \label{noise scheduler}
  \centering
  \adjustbox{width=0.75\textwidth}{
  \begin{tabular}{cccccc}
    \toprule
    \multirow{2}{*}{\textbf{Noise Scale}} & \multirow{2}{*}{\textbf{Metric}} & \multirow{2}{*}{\textbf{Noise Scheduler}} & \multicolumn{3}{c}{\textbf{Methods}} \\
    \cmidrule(r){4-6} & & & Clean & UDP & EUDP  \\
    \midrule
    \multirow{9}{*}{16/255} & FID$\uparrow$ & \multirow{3}{*}{\makecell[c]{$\beta_0=0.0001$\\$\beta_T=0.015$}} & 6.10 & 57.59 & 58.49 \\
    & Precision(\%)$\downarrow$ & & 69.52 & 36.40 & 30.59\\
    & Recall(\%)$\downarrow$ & & 53.25 & 24.74 & 24.12\\
    \cmidrule(r){2-6}
    & FID$\uparrow$ & \multirow{3}{*}{\makecell[c]{$\beta_0=0.0001$\\ $\beta_T=0.020$}} & 3.83 & 56.52 & 60.75 \\
    & Precision(\%)$\downarrow$ & & 71.77 & 36.04 & 30.88\\
    & Recall(\%)$\downarrow$ & & 54.13 & 29.58 & 23.81\\
    \cmidrule(r){2-6}
    & FID$\uparrow$ & \multirow{3}{*}{\makecell[c]{$\beta_0=0.0001$\\$\beta_T=0.025$}} & 5.84 & 54.89 & 60.47 \\
    & Precision(\%)$\downarrow$ & & 67.92 & 36.16 & 31.27\\
    & Recall(\%)$\downarrow$ & & 55.30 & 26.20 & 24.35\\
    \bottomrule
  \end{tabular}
  }
\end{table}

\subsection{More Visualization}
\label{sec:More Visualization}
\begin{figure}[htbp]
  \centering
  \subfigure[Clean data of Monet]{
    \includegraphics[width=0.31\textwidth]{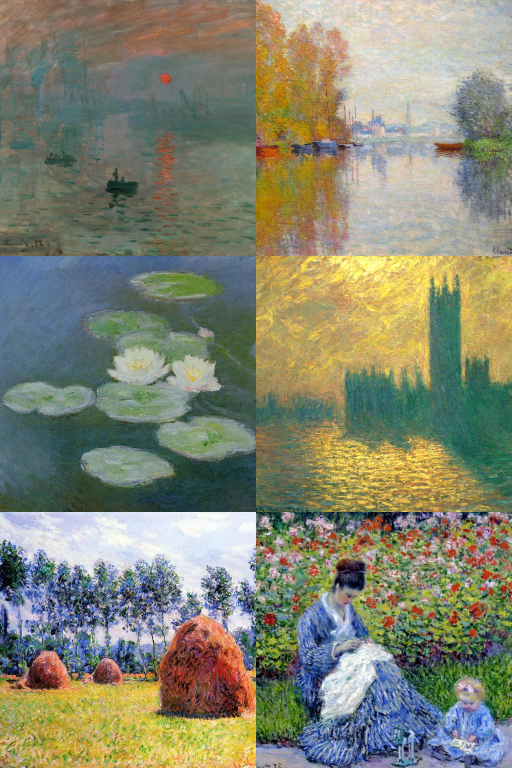}
    \label{fig:Monet}
  }
  \subfigure[Clean data of Picasso]{
    \includegraphics[width=0.31\textwidth]{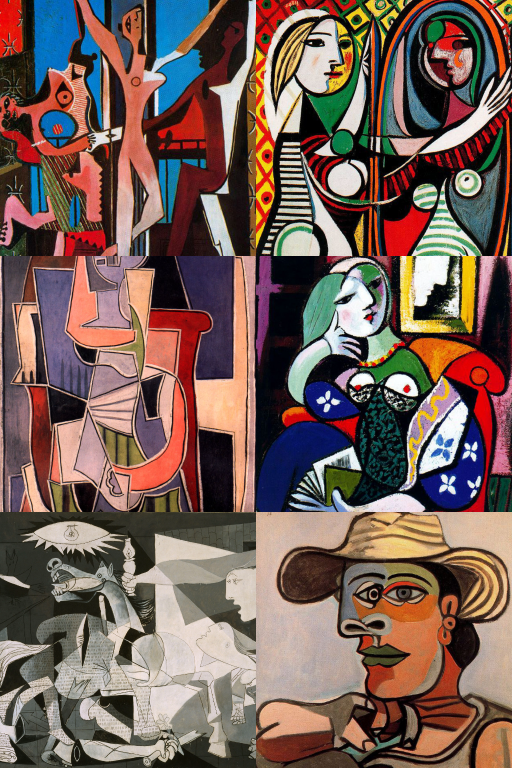}
    \label{fig:Picasso}
  }
  \subfigure[Clean data of van Gogh]{
    \includegraphics[width=0.31\textwidth]{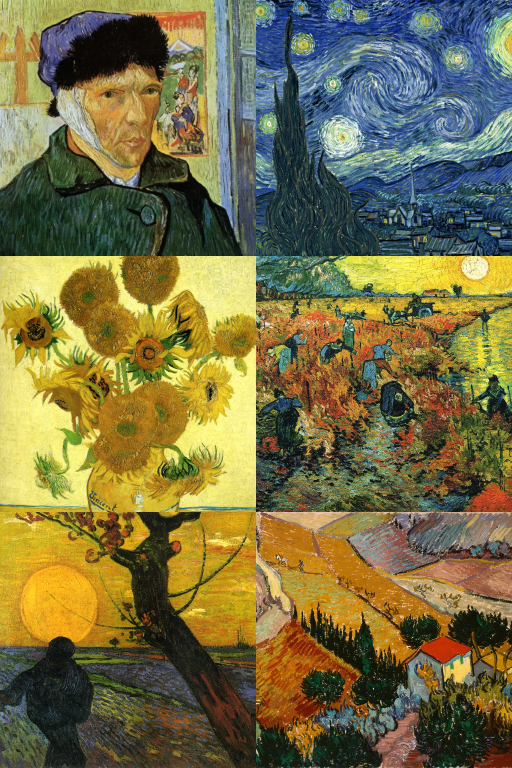}
    \label{fig:VG}
  }
  \subfigure[Protected data of Monet]{
    \includegraphics[width=0.31\textwidth]{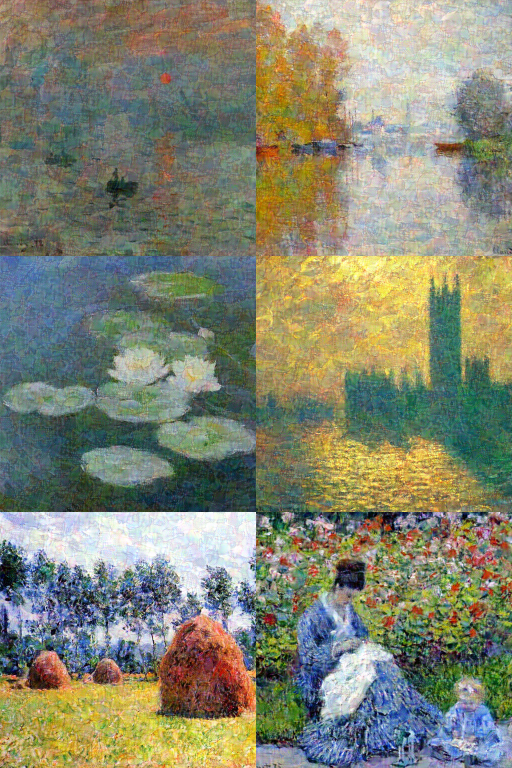}
    \label{fig:Monet_poisoned}
  }
  \subfigure[Protected data of Picasso]{
    \includegraphics[width=0.31\textwidth]{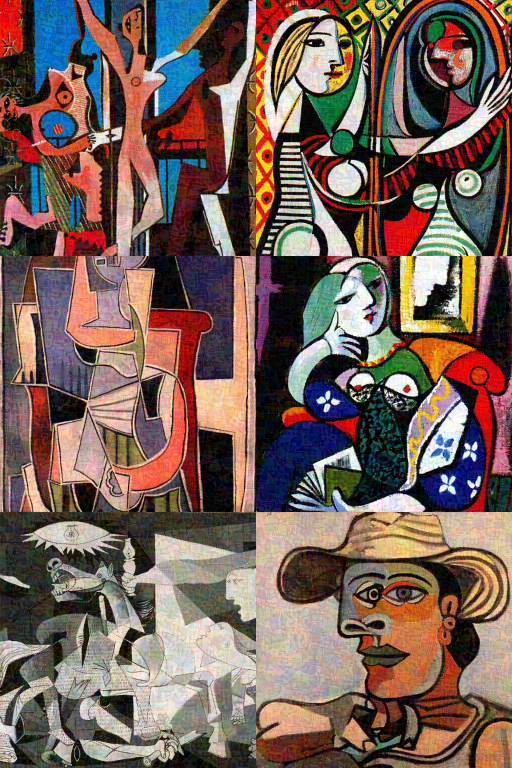}
    \label{fig:Picasso_poisoned}
  }
  \subfigure[Protected data of van Gogh]{
    \includegraphics[width=0.31\textwidth]{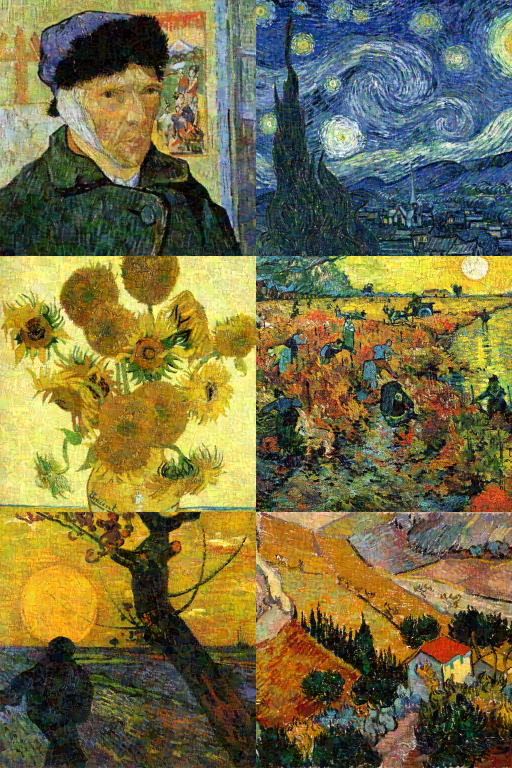}
    \label{fig:VG_poisoned}
  }
  \vspace{-3mm}
  \caption{\label{fig:artworks} Clean and protected artworks of different artists.}
\end{figure}

\begin{figure}[htbp]
  \centering
  \includegraphics[width=1\textwidth]{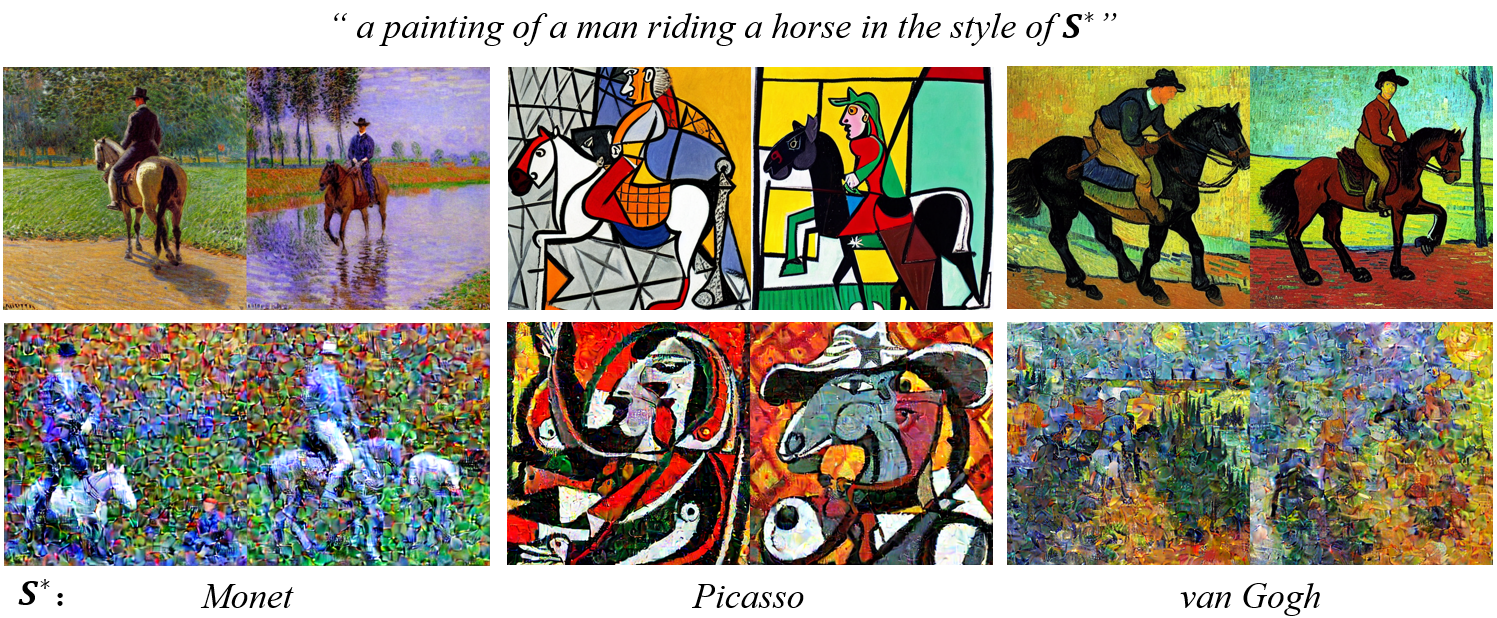}
  \vspace{-6mm}
  \caption{\label{fig:text2style_2} An example of text-to-image with specific styles. The first row: Generated images of Stable Diffusion fine-tuned on the clean dataset. The second row: Generated images of Stable Diffusion fine-tuned on the EUDP dataset.}
\end{figure}

\begin{figure}[htbp]
  \centering
  \subfigure[]{
    \includegraphics[width=1\textwidth]{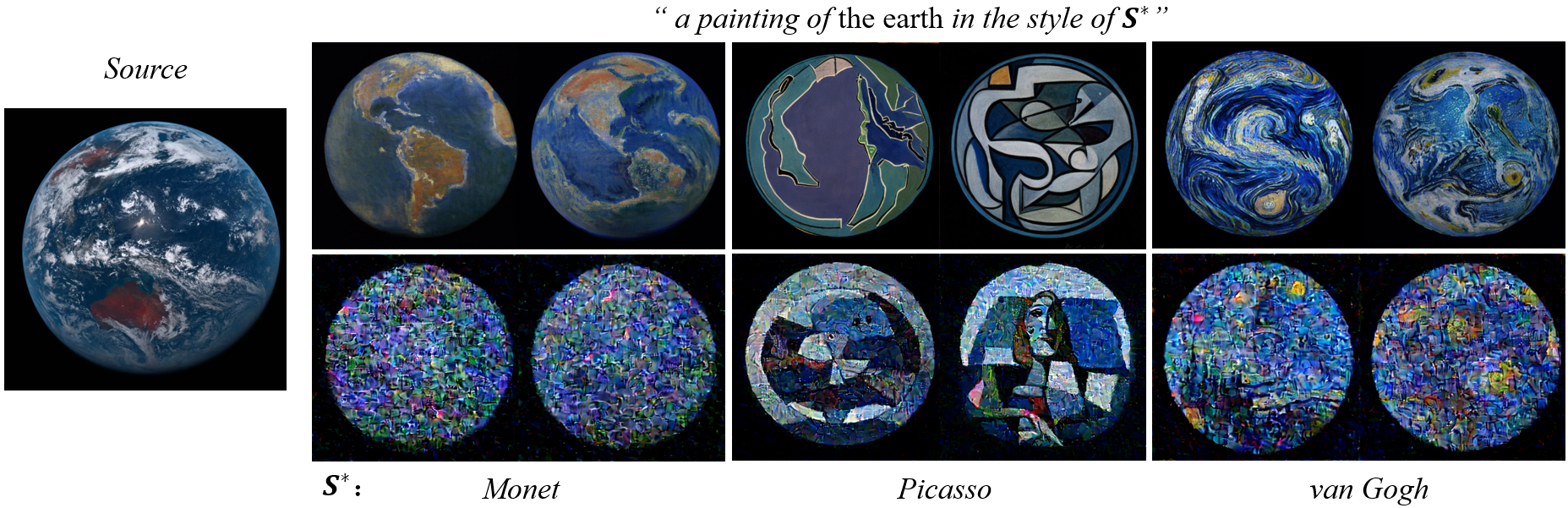}
    \label{fig:img2style_2}
  }
  \subfigure[]{
    \includegraphics[width=1\textwidth]{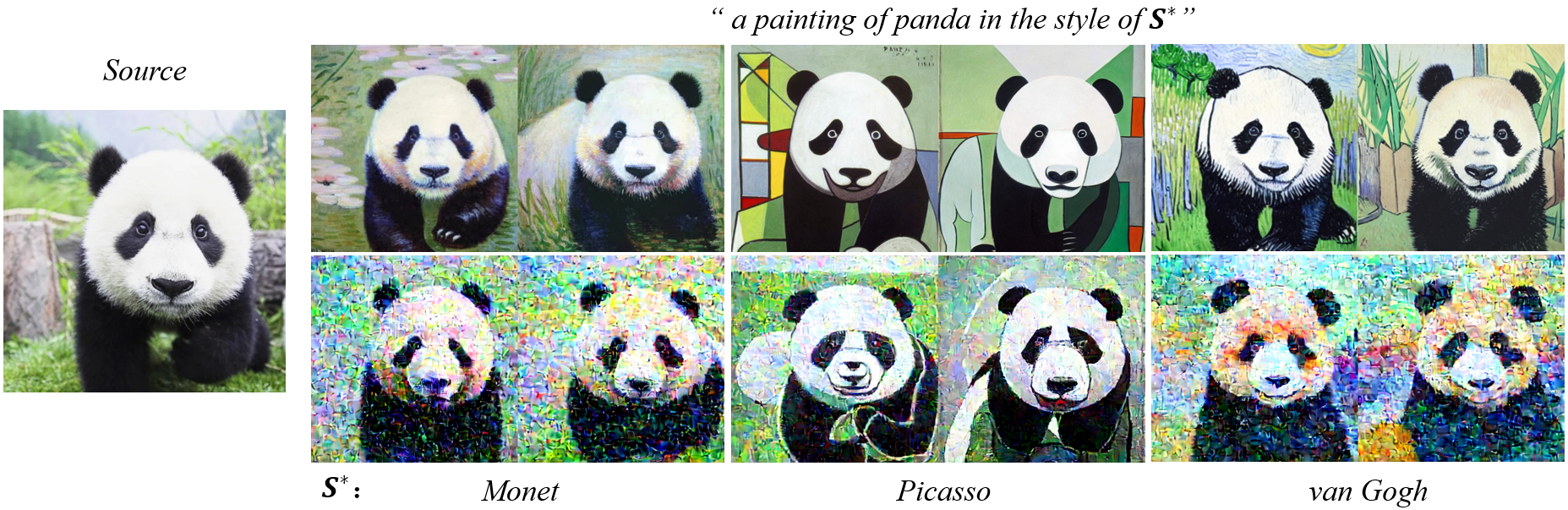}
    \label{fig:img2style_3}
  }
  \subfigure[]{
    \includegraphics[width=1\textwidth]{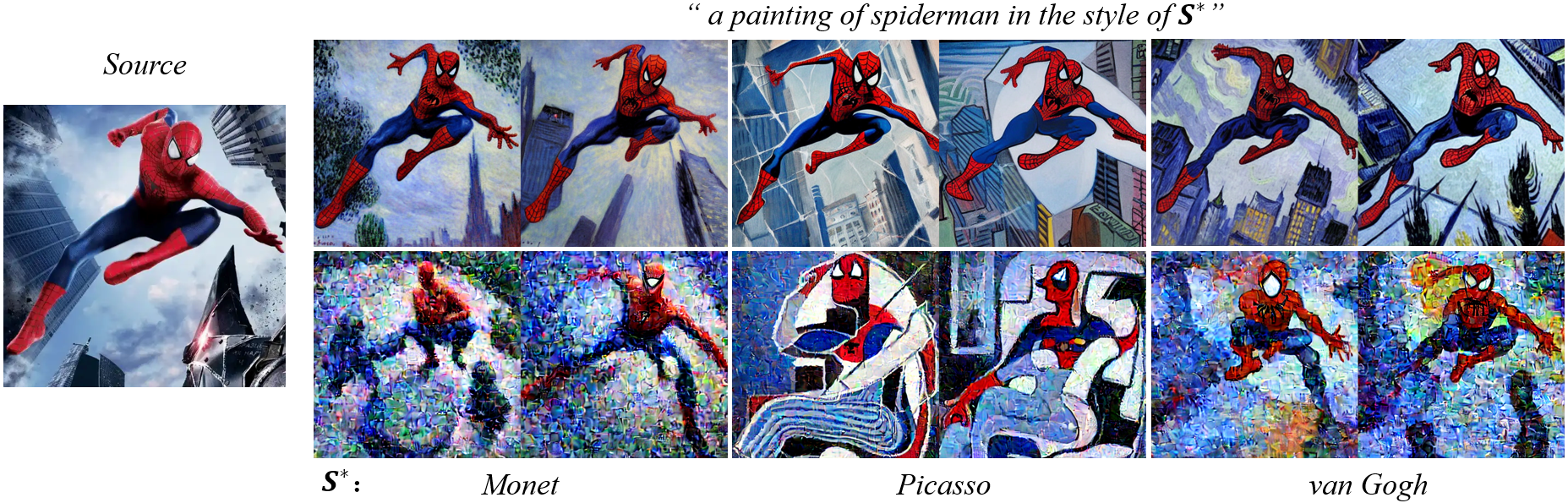}
    \label{fig:img2style_4}
  }
   \vspace{-3mm}
  \caption{\label{fig:img2style_n} Examples of style transfer. The first row: Generated images of Stable Diffusion fine-tuned on the clean dataset. The second row: Generated images of Stable Diffusion fine-tuned on the EUDP dataset.}
\end{figure}

Here we show some more visualization results of protected images, text-to-image, and style mimicry. We demonstrate the clean artworks and EUDP-protected artworks of different artists including Monet, Picasso, and van Gogh in Figure~\ref{fig:artworks}, text-to-image with specific styles in Figure~\ref{fig:text2style_2}, and style mimicry (image-to-image) in Figure~\ref{fig:img2style_n}, respectively.


\end{document}